\documentclass{article}

\usepackage{times}
\usepackage{graphicx} 
\usepackage{subfigure}

\usepackage{natbib}

\usepackage{algorithm}
\usepackage{algorithmic}

\usepackage{hyperref}

\usepackage[accepted]{icml2016_arxiv}

\usepackage[fleqn]{amsmath}
\usepackage{amssymb}
\usepackage{amsthm}
\usepackage{dsfont}
\usepackage{enumitem}

\newcommand{\Exp}{\mathds{E}}

\newcommand{\Prob}{\mathds{P}}
\newcommand{\Real}{\mathds{R}}
\newcommand{\Nat}{\mathbb{N}}

\newcommand{\so}{\succcurlyeq_{\rm so}}

\newcommand{\Zc}{\mathcal{Z}}

\newcommand{\Sc}{\mathcal{S}}

\newcommand{\Mc}{\mathcal{M}}

\newcommand{\Bc}{\mathcal{B}}
\newcommand{\Hc}{\mathcal{H}}

\newcommand{\CASE}[1]{\STATE \textbf{case} #1\textbf{:} \begin{ALC@g}}
\newcommand{\ENDCASE}{\end{ALC@g}}

\newcommand{\DEFAULT}{\STATE \textbf{default:} \begin{ALC@g}}
\newcommand{\ENDDEFAULT}{\end{ALC@g}}
\newcommand{\DEFAULTLINE}[1]{\STATE \textbf{default:} }

\newtheorem{theorem}{Theorem}
\newtheorem{lemma}{Lemma}

\newtheorem{conjecture}{Conjecture}

\newtheorem{example}{Example}

\newtheorem{definition}{Definition}

\newcommand{\state}{\mathcal{S}}

\newcommand{\action}{\mathcal{A}}

\newcommand{\mdp}{\mathcal{M}}

\newcommand{\reg}{\mathrm{Regret}}

\DeclareMathOperator*{\argmax}{arg\,max}
\DeclareMathOperator*{\argmin}{arg\,min}

\newcommand{\E}{\mathds{E}}
\newcommand{\Ind}{\mathds{1}}

\icmltitlerunning{Generalization and Exploration via Randomized Value Functions}

\begin{document}

\twocolumn[
\icmltitle{Generalization and Exploration via Randomized Value Functions}

\icmlauthor{Ian Osband}{iosband@stanford.edu}
\icmlauthor{Benjamin Van Roy}{bvr@stanford.edu}
\icmlauthor{Zheng Wen}{zhengwen207@gmail.com}
\icmladdress{Stanford University}

\vskip 0.3in
]

\begin{abstract}
We propose randomized least-squares value iteration (RLSVI) -- a new reinforcement learning
algorithm designed to explore and generalize efficiently via linearly parameterized value functions.
We explain why versions of least-squares value iteration
that use Boltzmann or $\epsilon$-greedy exploration can be highly inefficient, and we present computational results
that demonstrate dramatic efficiency gains enjoyed by RLSVI.
Further, we establish an upper bound on the expected regret of RLSVI that demonstrates near-optimality
in a {\it tabula rasa} learning context.  More broadly, our results suggest that randomized value functions
offer a promising approach to tackling a critical challenge in reinforcement learning: synthesizing efficient
exploration and effective generalization.
\end{abstract}

\vspace{-4mm}
\section{Introduction}
\label{sec:introduction}

The design of reinforcement learning (RL) algorithms that explore
intractably large state-action spaces efficiently remains an important challenge.
In this paper, we propose randomized
least-squares value iteration (RLSVI), which generalizes using a linearly parameterized value function.  Prior RL algorithms
that generalize in this way require, in the worst case,
learning times exponential in the number of model parameters and/or the planning horizon.  RLSVI
aims to overcome these inefficiencies.

RLSVI operates in a manner similar to least-squares value iteration (LSVI) and also shares much of the spirit of other closely related approaches such as
TD, LSTD, and SARSA (see, e.g., \cite{Sutton1998,2010Szepesvari}).  What fundamentally distinguishes RLSVI is that the algorithm explores through randomly sampling
statistically plausible value functions, whereas the aforementioned alternatives are typically applied in conjunction with action-dithering schemes
such as Boltzmann or $\epsilon$-greedy exploration, which lead to highly inefficient learning.  The concept of exploring by sampling statistically
plausible value functions is broader than any specific algorithm, and beyond our proposal and study of RLSVI. We view an important role of this paper is to establish this broad concept as a promising approach to tackling a critical challenge in RL: synthesizing efficient
exploration and effective generalization.

We will present computational results comparing RLSVI to LSVI with action-dithering schemes.  In our case studies, these
algorithms generalize using identical linearly parameterized value functions but are distinguished by how they explore.
The results demonstrate that RLSVI enjoys dramatic efficiency gains.
Further, we establish a bound on the expected regret for an episodic {\it tabula rasa} learning context.
Our bound is $\tilde{O}(\sqrt{H^3 S A T})$, where $S$ and $A$ denote the cardinalities of the
state and action spaces, $T$ denotes time elapsed, and $H$ denotes the episode duration.
This matches the worst case lower bound for this problem up to logarithmic factors \cite{Jaksch2010}.
It is interesting to contrast this against known $\tilde{O}(\sqrt{H^3 S^2 A T})$ bounds for other
provably efficient {\it tabula rasa} RL algorithms
(e.g., UCRL2 \cite{Jaksch2010}) adapted to this context.
To our knowledge, our results establish RLSVI as the
first RL algorithm that is provably efficient in a {\it tabula rasa} context and also demonstrates efficiency when generalizing via linearly parameterized value functions.

There is a sizable literature on RL algorithms that are provably efficient in {\it tabula rasa} contexts
\citep{Brafman2002, Kakade2003,  Kearns1999, Lattimore2013, Ortner2012,  Osband2013, Strehl2006}.  The literature
on RL algorithms that generalize and explore in a provably efficient manner is sparser.
There is work on model-based RL algorithms \citep{Abbasi-Yadkori2011,osband2014model,osband2014near},
which apply to specific model classes and are computationally intractable.  Value function generalization
approaches have the potential to overcome those computational challenges and offer practical means
for synthesizing efficient exploration and effective generalization.
A relevant line of work establishes that efficient RL with value function generalization reduces
to efficient KWIK online regression \citep{Li2010,LiLW08}.  However,
it is not known whether the KWIK online regression problem can be solved efficiently.  In terms of concrete algorithms, there is
optimistic constraint propagation (OCP) \citep{WenVanroy13}, a provably efficient RL
algorithm for exploration and value function generalization in deterministic systems, and C-PACE \citep{pazis2013pac},
a provably efficient RL algorithm that generalizes using interpolative representations.
These contributions represent important developments, but OCP is not suitable for stochastic systems and is highly sensitive to
model mis-specification, and generalizing effectively in high-dimensional state spaces calls for methods that extrapolate.
RLSVI advances this research agenda, leveraging randomized value functions to explore efficiently with linearly parameterized
value functions.  The only other work we know of involving exploration through random sampling of value functions is \citep{DeardenFR98}.
That work proposed an algorithm for {\it tabula rasa} learning; the algorithm does not generalize
over the state-action space.

\section{Episodic reinforcement learning}
\label{sec:problem_formulation}

{\medmuskip=1mu
\thinmuskip=1mu
\thickmuskip=1mu
A finite-horizon MDP $\mdp=\left( \state, \action, H, P, R, \pi \right)$, where $\state$ is a finite state space, $\action$ is a finite action space, $H$ is the number of periods, $P$ encodes transition probabilities, $R$ encodes reward distributions, and $\pi$ is a state distribution.
In each episode, the initial state $s_0$ is sampled from $\pi$, and, in period $h=0,1, \cdots, H-1$, if the state is $s_h$ and an action $a_h$ is selected then a next state $s_{h+1}$ is sampled from $P_h(\cdot | s_h,a_h)$ and a reward $r_h$ is sampled from $R_h\left(\cdot \middle | s_h,a_h,s_{h+1} \right)$.
The episode terminates when state $s_H$ is reached and a terminal reward is sampled from $R_H\left(\cdot \middle | s_H \right)$.
}

To represent the history of actions and observations over multiple episodes, we will often index variables by both episode and period.
For example, $s_{lh}$, $a_{lh}$ and $r_{lh}$ respectively denote the state, action, and reward observed during period $h$
in episode $l$.

A policy $\mu=\left ( \mu_0, \mu_1, \cdots, \mu_{H-1} \right)$ is a sequence of functions, each mapping $\state$ to $\action$.
For each policy $\mu$, we define a value function for $h = 0,..,H$:
{\small \medmuskip=0mu
\thinmuskip=0mu
\thickmuskip=0mu
$$V_h^\mu(s):=\E_\mdp\left[ \textstyle \sum_{\tau=h}^H r_\tau \middle | s_h =s, a_\tau = \mu_\tau(s_\tau) \text{ for } \tau=h,..,H-1 \right]$$ }
The optimal value function is defined by
$V^*_h(s) = \sup_\mu V^\mu_h(s)$.
A policy $\mu^*$ is said to be optimal if $V^{\mu^*} = V^*$.
It is also useful to define a state-action optimal value function for $h = 0,..,H-1$:
\begin{equation*}
Q^*_h(s,a) := \E_\mdp\left[ r_h + V^*_{h+1} (s_{h+1}) \middle | s_h=s, a_h=a \right]
\end{equation*}
{\small \medmuskip=0mu
\thinmuskip=0mu
\thickmuskip=0mu
A policy $\mu^*$ is optimal $\iff$
$\mu_h^*(s) \in \argmax_{\alpha \in \action} Q^*_h(s,\alpha)$, $\forall s,h$.

A reinforcement learning algorithm generates each action $a_{lh}$ based on observations made up
to period $h$ of episode $l$.  Over each episode, the algorithm realizes reward
$\sum_{h=0}^H r_{lh}$.  One way to quantify the performance of a reinforcement learning algorithm is in terms of the
\textit{expected cumulative regret} over $L$ episodes, or time $T = LH$, defined by
}
\begin{equation*}
{\small
\medmuskip=0mu
\thinmuskip=0mu
\thickmuskip=0mu
\reg(T, \mdp)=\textstyle \sum_{l=0}^{T/H-1} \E_\mdp\left[ V^*_0(s_{l0}) - \textstyle \sum_{h=0}^H r_{lh} \right].}
\label{eqn:regret}
\end{equation*}

Consider a scenario in which the agent models that,
for each $h$, $Q^*_h \in \mathrm{span}\left[ \Phi_h \right]$ for some $\Phi_h \in \Real^{\state \action \times K}$. With some abuse of notation, we use $\state$ and $\action$ to denote the cardinalities of the state and action spaces.
We refer this matrix $\Phi_h$ as a \textit{generalization matrix} and use $\Phi_h(s,a)$ to denote the row of matrix $\Phi_h$ associated with state-action pair $(s,a)$.
For $k=1,2,\cdots, K$, we write the $k$th column of $\Phi_h$ as $\phi_{hk}$ and refer to $\phi_{hk}$ as a basis function.
We refer to contexts where the agent's belief is correct as {\it coherent learning}, and refer the alternative as {\it agnostic learning}.

\section{The problem with dithering for exploration}
\label{sec:inefficient_schemes}

LSVI can be applied at each episode to estimate the optimal value function $Q^*$ from data gathered over previous episodes.
To form an RL algorithm based on LSVI, we must specify how the agent selects actions.
The most common scheme is to selectively take actions at random, we call this approach dithering.
Appendix \ref{sec:Boltzmann_epsilon} presents RL algorithms resulting from combining LSVI with the most common schemes of $\epsilon$-greedy or Boltzmann exploration.

The literature on efficient RL shows that these dithering schemes can lead to regret that grows exponentially in $H$ and/or $\Sc$ \cite{Kearns2002,Brafman2002,Kakade2003}.
Provably efficient exploration schemes in RL require that exploration is directed towards potentially informative state-action pairs and consistent over multiple timesteps.
This literature provides several more intelligent exploration schemes that are provably efficient, but most only apply to {\it tabula rasa} RL, where little prior information is available and learning is considered efficient even if the time required scales with the cardinality of the state-action space.
In a sense, RLSVI represents a synthesis of ideas from efficient {\it tabula rasa} reinforcement learning and value function generalization methods.

\begin{figure}[h!]
\vspace{-2mm}
\centering
\includegraphics[width=0.99\linewidth]{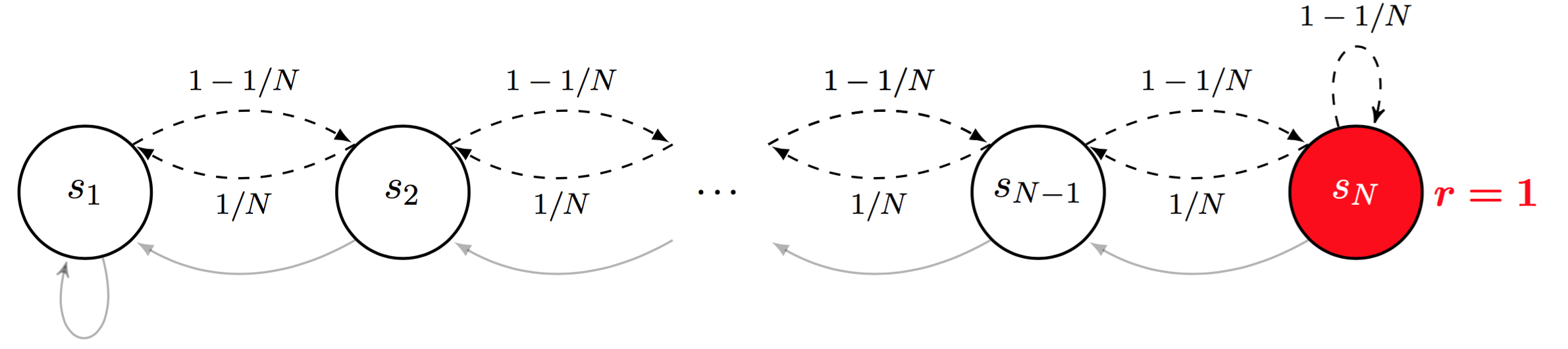}
\vspace{-3mm}
\caption{\small An MDP where dithering schemes are highly inefficient.}
\label{fig:example1}
\end{figure}

To motivate some of the benefits of RLSVI, in Figure \ref{fig:example1} we provide a simple example that highlights the failings of dithering methods.
In this setting LSVI with Boltzmann or $\epsilon$-greedy exploration requires exponentially many episodes to learn an optimal policy, even in a coherent learning context and even with a small number of basis functions.

This environment is made up of a long chain of states $\Sc = \{1,..,N\}$.
Each step the agent can transition left or right.
Actions left are deterministic, but actions right only succeed with probability $1-1/N$, otherwise they go left.
All states have zero reward except for the far right $N$ which gives a reward of $1$.
Each episode is of length $H=N-1$ and the agent will begin each episode at state $1$.
The optimal policy is to go right at every step to receive an expected reward of $p^\ast=(1-\frac{1}{N})^{N-1}$ each episode, all other policies give no reward.
Example \ref{example:example1} establishes that, for any choice of basis function, LSVI with any $\epsilon$-greedy or Boltzmann exploration will lead to regret that grows exponentially in $\Sc$.
A similar result holds for policy gradient algorithms.

\vspace{-2mm}

\begin{example}
\label{example:example1}
Let $l^*$ be the first episode during which state $N$ is visited.
It is easy to see that $\theta_{lh}=0$ for all $h$ and all $l < l^*$.
Furthermore, with either $\epsilon$-greedy or Boltzmann exploration, actions are sampled uniformly at random over episodes $l < l^*$.
Thus, in any episode $l < l^*$, the red node will be reached with probability
$p^\ast 2^{- (\state-1)}=p^\ast 2^{-H}$.
It follows that $E[l^*] \geq 2^{\state-1}-1$ and
$\liminf_{T \rightarrow \infty} \reg(T,\mdp) \geq 2^{\state-1} - 1$.
\end{example}

\section{Randomized value functions}
\label{sec:rlsvi}

We now consider an
alternative approach to exploration that involves randomly sampling value functions rather than actions.  As a specific
scheme of this kind, we propose randomized least-squares value iteration (RLSVI), which we present as
Algorithm \ref{alg:rlsvi}.\footnote{Note that when $l=0$, both $A$ and $b$ are empty, hence, we set $\tilde{\theta}_{l0}= \tilde{\theta}_{l1}= \cdots = \tilde{\theta}_{l,H-1}=0$.}
To obtain an RL algorithm, we simply select greedy actions in each episode, as specified in Algorithm \ref{alg:greedy_original}.

The manner in which RLSVI explores is inspired by Thompson sampling \cite{Thompson1933}, which has been shown to explore efficiently across a very general class of online optimization problems \cite{Russo2013b,Russo2013}.
In Thompson sampling, the agent samples from a posterior distribution over models, and selects the action that optimizes the sampled model.
RLSVI similarly samples from a distribution over plausible value functions and selects actions that optimize resulting samples.
This distribution can be thought of as an approximation to a posterior distribution over value functions.
RLSVI bears a close connection to PSRL \cite{Osband2013}, which maintains and samples from a posterior distribution over MDPs and is a direct application of Thompson sampling to RL.
PSRL satisfies regret bounds that scale with the dimensionality, rather than the cardinality, of the underlying MDP \cite{osband2014near,osband2014model}.
However, PSRL does not accommodate value function generalization without MDP planning, a feature that we expect to be of great practical importance.

{
\medmuskip=1mu
\thinmuskip=0mu
\thickmuskip=2mu

\small
\begin{algorithm}[!h]
\caption{Randomized Least-Squares Value Iteration}
\label{alg: RLSVI}
\textbf{Input:} Data $\Phi_0(s_{i0},a_{i0}),r_{i0},..,\Phi_{H-1}(s_{iH-1}, a_{iH-1}), r_{iH} : i < L$,
Parameters $\lambda > 0, \  \sigma > 0$ \\
\textbf{Output:} $\tilde{\theta}_{l0}, .., \tilde{\theta}_{l,H-1}$
\begin{algorithmic}[1]
\label{alg:rlsvi}
\FOR{$h=H-1, .., 1, 0$}
\STATE Generate regression problem $A \in \Real^{l \times K}$, $b \in \Real^l$:
\small
\begin{equation*}
    \begin{aligned}
        & A \leftarrow \left[\begin{array}{c} \Phi_h(s_{0 h}, a_{0 h}) \\ \vdots \\ \Phi_h(s_{l-1,h}, a_{l-1,h}) \end{array}\right] \\
        & b_i \leftarrow \left\{\begin{array}{ll}
    r_{ih} + \max_\alpha  \left( \Phi_{h+1} \tilde{\theta}_{l, h+1} \right) \left( s_{i,h+1} , \alpha \right) & \text{if } h < H-1 \\
    r_{ih} + r_{i,h+1} \qquad & \text{if } h = H - 1
    \end{array}\right.
    \end{aligned}
\end{equation*}
\normalsize
\STATE Bayesian linear regression for the value function
\small
\begin{equation*}
    \begin{aligned}
        & \overline{\theta}_{lh} \leftarrow \frac{1}{\sigma^2} \left(\frac{1}{\sigma^2} A^\top A + \lambda I \right)^{-1} A^{\top} b \\
        & \Sigma_{lh} \leftarrow \left(\frac{1}{\sigma^2} A^\top A + \lambda I \right)^{-1}
    \end{aligned}
\end{equation*}
\normalsize
\STATE Sample $\tilde{\theta}_{lh} \sim N(\overline{\theta}_{lh},\Sigma_{lh})$ from Gaussian posterior
\ENDFOR
\end{algorithmic}
\end{algorithm}
\normalsize

\small
\begin{algorithm}[!h]
\caption{RLSVI with greedy action}
\textbf{Input:} Features $\Phi_0,..,\Phi_{H-1}$; $\sigma >0, \ \lambda >0$ \\
\vspace{-4mm}
\begin{algorithmic}[1]
\label{alg:greedy_original}
\FOR{$l=0,1,..$}
\STATE Compute $\tilde{\theta}_{l 0}, .., \tilde{\theta}_{l, H-1}$ using Algorithm
\ref{alg:rlsvi}
\STATE Observe $s_{l 0}$
\FOR{$h=0,.., H-1$}
\STATE Sample $a_{l h} \in \argmax_{\alpha \in \action} \left( \Phi_h \tilde{\theta}_{lh} \right) \left( s_{lh}, \alpha \right)$
\STATE Observe $r_{lh}$ and $s_{l,h+1}$
\ENDFOR
\STATE Observe $r_{lH}$
\ENDFOR
\end{algorithmic}
\end{algorithm}
\normalsize
}

\vspace{-5mm}
\section{Provably efficient tabular learning}
\label{sec: proof tabular}

RLSVI is an algorithm designed for efficient exploration in large MDPs with linear value function generalization.
So far, there are no algorithms with analytical regret bounds in this setting.
In fact, most common  methods are provably \textit{inefficient}, as demonstrated in Example \ref{example:example1}, regardless of the choice of basis function.
In this section we will establish an expected regret bound for RLSVI in a tabular setting without generalization where the basis functions $\Phi_h = I$.

The bound is on an expectation with respect to a probability space $(\Omega, \mathcal{F}, \mathbb{P})$.
We define the MDP $\mdp = (\state, \action, H, P, R, \pi)$ and all other random variables we will consider with respect to this probability space.
We assume that $\state$, $\action$, $H$, and $\pi$, are deterministic and that $R$ and $P$ are drawn from a prior distribution.
We will assume that rewards $R(s,a,h)$ are drawn from independent Dirichlet $\alpha^R(s,a,h) \in \Real_+^2$ with values on $\{-1, 0\}$ and transitions Dirichlet $\alpha^P(s,a,h) \in \Real_+^S$.
Analytical techniques exist to extend similar results to general bounded distributions; see, for example \cite{agrawal2012further}.


\begin{theorem}
\label{thm: regret}
{
\medmuskip=0mu
\thinmuskip=0mu
\thickmuskip=0mu
If Algorithm \ref{alg: RLSVI} is executed with $\Phi_h = I$ for $h=0,..,H-1$,
$\lambda \ge \max_{(s,a,h)} \left(\Ind^T\alpha^R(s,a,h) + \Ind^T\alpha^P(s,a,h)\right)$
and $\sigma \ge \sqrt{H^2 + 1}$, then:
}
\begin{equation}
    \Exp \left[{\rm Regret}(T, \Mc) \right]
        \le \tilde{O}\left(\sqrt{H^3 SAT}\right)
\end{equation}
\end{theorem}

Surprisingly, these scalings better state of the art optimistic algorithms specifically designed for efficient analysis which would admit $\tilde{O}(\sqrt{H^3 S^2 A T})$ regret \cite{Jaksch2010}.
This is an important result since it demonstrates that RLSVI can be provably-efficient, in contrast to popular dithering approaches such as $\epsilon$-greedy which are provably \textit{inefficient}.

\subsection{Preliminaries}

Central to our analysis is the notion of stochastic optimism, which induces a partial ordering among random variables.
\begin{definition}
\label{def:stochastic optimism}
For any $X$ and $Y$ real-valued random variables we say that $X$ is stochastically optimistic for $Y$ if and only if for any
{\medmuskip=1mu
\thinmuskip=1mu
\thickmuskip=1mu
$u:\Real \rightarrow \Real$}
convex and increasing
$$\E[u(X)] \geq \E[u(Y)].$$
We will use the notation $X \so Y$ to express this relation.
\end{definition}
\vspace{-0.1in}

It is worth noting that stochastic optimism is closely connected with second-order stochastic dominance: $X \so Y$ if and only if $-Y$ second-order stochastically dominates $-X$ \cite{hadar1969rules}.
We repoduce the following result which establishes such a relation involving Gaussian and Dirichlet random variables in Appendix \ref{app: Gauss Dirichlet}.
\begin{lemma}
\label{lem:Gaussian-Dirichlet}
For all $V \in [0,1]^N$ and $\alpha \in [0,\infty)^N$ with $\alpha^T \Ind \ge 2$,
if $X \sim N(\alpha^\top V / \alpha^\top {\bf 1}, 1/ \alpha^\top {\bf 1})$
and $Y = P^T V$ for $P \sim {\rm Dirichlet}(\alpha)$
then $X \so Y$.
\end{lemma}
\vspace{-0.1in}

\subsection{Proof sketch}
Let $\tilde{Q}_h^l= \Phi_h \tilde{\theta}_{lh}$ and $\tilde{\mu}_l$ denote the value function and policy generated by RLSVI for episode $l$ and let $\tilde{V}_h^l(s) = \max_a \tilde{Q}_h^l(s,a)$.
We can decompose the per-episode regret
\begin{equation*}
{\small \medmuskip=1mu
\thinmuskip=0mu
\thickmuskip=1mu
V^*_0(s_{l0}) - V^{\tilde{\mu}_l}_0(s_{l0}) = \underbrace{\tilde{V}^l_0(s_{l0}) - V^{\tilde{\mu}_l}_0(s_{l0})}_{\Delta^{\rm conc}_l} + \underbrace{V^*_0(s_{l0}) - \tilde{V}^l_0(s_{l0})}_{\Delta^{\rm opt}_l}.
}
\end{equation*}
We will bound this regret by first showing that RLSVI generates optimistic estimates of $V^*$, so that $\Delta^{\rm opt}_l$ has nonpositive expectation for any history $\Hc_l$ available prior to episode $l$.
The remaining term $\Delta^{\rm conc}_l$ vanishes as estimates generated by RLSVI concentrate around $V^*$.

\begin{lemma}
\label{lem: optimistic Q}
Conditional on any data $\Hc$, the Q-values generated by RLSVI are stochastically optimistic for the true Q-values
$\tilde{Q}^l_{h}(s,a) \so Q^*_{h}(s,a)$ for all $s,a,h$.
\end{lemma}
\vspace{-4mm}
\begin{proof}
Fix any data $\Hc_l$ available and use backwards induction on $h=H-1,..,1$.
For any $(s,a,h)$ we write $n(s,a,h)$ for the amount of visits to that datapoint in $\Hc_l$.
We will write $\hat{R}(s,a,h), \hat{P}(s,a,h)$ for the empirical mean reward and mean transitions based upon the data $\Hc_l$.
We can now write the posterior mean rewards and transitions:
$$ \overline{R}(s,a,h) | \Hc_l = \frac{-1 \times \alpha^R_1(s,a,h) + n(s,a,h) \hat{R}(s,a,h)}{\Ind^T \alpha^R(s,a,h) + n(s,a,h)}$$
$$ \overline{P}(s,a,h) | \Hc_l = \frac{\alpha^P(s,a,h) + n(s,a,h) \hat{P}(s,a,h)}{\Ind^T \alpha^P(s,a,h) + n(s,a,h)}$$

Now, using $\Phi_h = I$ for all $(s,a,h)$ we can write the RLSVI updates in similar form.
Note that, $\Sigma_{lh}$ is diagonal with each diagonal entry equal to $\sigma^2 / (n(s,a,h) + \lambda \sigma^2)$.
In the case of $h=H-1$
$$ \overline{\theta}^l_{H-1}(s,a) = \frac{n(s,a,H-1)\hat{R}(s,a,H-1)}{n(s,a,H-1) + \lambda \sigma^2} $$
Using the relation that $\hat{R} \ge \overline{R}$ Lemma \ref{lem:Gaussian-Dirichlet} means that
$$ N(\overline{\theta}^l_{H-1}(s,a), \frac{1}{n(s,a,h) + \Ind^T \alpha^R(s,a,h)}) \so R_{H-1} | \Hc_l .$$
Therefore, choosing $\lambda > \max_{s,a,h} \Ind^T \alpha^R(s,a,h)$ and $\sigma > 1$, we must satisfy the lemma for all $s,a$ and $h=H-1$.

For the inductive step we assume that the result holds for all $s, a $ and $j > h$, we now want to prove the result for all $(s,a)$ at timestep $h$.
Once again, we can express $\overline{\theta}^l_h(s,a)$ in closed form.
$$ \overline{\theta}^l_{h}(s,a) =
\frac{n(s,a,h) \left(\hat{R}(s,a,h) + \hat{P}(s,a,h)^T \tilde{V}^l_{h+1} \right)}{n(s,a,h) + \lambda \sigma^2}$$
To simplify notation we omit the arguments $(s,a,h)$ where they should be obvious from context.
The posterior mean estimate for the next step value $V^*_h$, conditional on $\Hc_l$:
$$ \Exp[Q^*_h(s,a) | \Hc_l ] = \overline{R} + \overline{P}^T V^*_{h+1} \le \frac{n (\hat{R} + \hat{P}^T V^*_{h+1})}{n + \lambda \sigma^2}.$$
As long as $\lambda > \Ind^T \alpha^R + \Ind^T(\alpha^P)$ and $\sigma^2 > H^2$.
By our induction process $\tilde{V}^l_{h+1} \so V^*_{h+1}$ so that
$$\Exp[Q^*_h(s,a) | \Hc_l] \le
    \Exp \left[ \frac{n (\hat{R} + \hat{P}^T \tilde{V}^l_{h+1})}{n + \lambda \sigma^2} \mid \Hc_l\right].$$
We can conclude by Lemma \ref{lem:Gaussian-Dirichlet} and noting that the noise from rewards is dominated by $N(0,1)$ and the noise from transitions is dominated by $N(0, H^2)$.
This requires that $\sigma^2 \ge H^2 + 1$.
\end{proof}

Lemma \ref{lem: optimistic Q} means RLSVI generates stochastically optimistic Q-values for any history $\Hc_l$.
All that remains is to prove the remaining estimates $\Exp[ \Delta^{\rm conc}_l | \Hc_l ]$ concentrate around the true values with data.
Intuitively this should be clear, since the size of the Gaussian perturbations decreases as more data is gathered.
In the remainder of this section we will sketch this result.

The concentration error
$ \Delta^{\rm conc}_l = \tilde{V}^l_0(s_{l0}) - V^{\tilde{\mu}_l}_0(s_{l0})$.
We decompose the value estimate $\tilde{V}^l_0$ explicitly:
\begin{eqnarray*}
    \tilde{V}^l_0(s_{l0}) &=&
        \frac{n (\hat{R} + \hat{P}^T \tilde{V}^l_{h+1})}{n + \lambda \sigma^2} + w^\sigma \\
        &=& \overline{R} + \overline{P}^T \tilde{V}^l_{h+1} + b^R + b^P + w^\sigma_0
\end{eqnarray*}
where $w^\sigma$ is the Gaussian noise from RLSVI and $b^R=b^R(s_{l0}, a_{l0} 0), b^P=b^P(s_{l0}, a_{l0} 0)$ are optimistic bias terms for RLSVI.
These terms emerge since RLSVI shrinks estimates towards zero rather than the Dirichlet prior for rewards and transitions.

Next we note that, conditional on $\Hc_l$ we can rewrite $\overline{P}^T \tilde{V}^l_{h+1} = \tilde{V}^l_{h+1}(s') + d_h$ where $s' \sim P^*(s,a,h)$ and $d_h$ is some martingale difference.
This allows us to decompose the error in our policy to the estimation error of the states and actions we actually visit.
We also note that, conditional on the data $\Hc_l$ the true MDP is independent of the sampling process of RLSVI.
This means that:
\begin{eqnarray*}
    \Exp[V^{\tilde{\mu}_l}_0(s_{l0}) | \Hc_l]
        &=& \overline{R} + \overline{P}^T V^{\tilde{\mu}_l}_{h+1}.
\end{eqnarray*}
Once again, we can replace this transition term with a single sample $s' \sim P^*(s,a,h)$ and a martingale difference.
Combining these observations allows us to reduce the concentration error
\begin{equation*}
    \begin{aligned}
        & \Exp[\tilde{V}^l_0(s_{l0}) - V^{\tilde{\mu}_l}_0(s_{l0}) | \Hc_l]
        = \\
        & \sum_{h=0}^{H-1} \left\{ b^R(s_{lh}, a_{lh}, h) + b^P(s_{lh}, a_{lh}, h) + w^\sigma_h \right\}.
    \end{aligned}
\end{equation*}
\vspace{-6mm}

We can even write explicit expressions for $b^R, b^P$ and $w^\sigma$.
\vspace{-3mm}
{\small
\begin{equation*}
    \begin{aligned}
        & b^R(s,a,h) = \frac{n \hat{R} }{n + \lambda \sigma^2 }
            - \frac{n \hat{R} - \alpha_1^R}{n + \Ind^T\alpha^R}  \\
        & b^P(s,a,h) = \frac{n \hat{P}^T \tilde{V}^l_{h+1} }{n + \lambda \sigma^2 } - \frac{(n \hat{P} + \alpha^P)^T \tilde{V}^l_{h+1}}{n + \Ind^T\alpha^P} \\
        & w^\sigma_h \sim N\left(0, \frac{\sigma^2}{n + \lambda \sigma^2}\right)
    \end{aligned}
\end{equation*}
}

\vspace{-3mm}

{
\medmuskip=0mu
\thinmuskip=0mu
\thickmuskip=0mu
The final details for this proof are technical but the argument is simple.
We let $\lambda = \Ind^T \alpha^R + \Ind^T \alpha^P$ and $\sigma = \sqrt{H^2 +1}$.
Up to $\tilde{O}$ notation $b^R \simeq \frac{\alpha_1^R}{n + \Ind^T \alpha^P}$, $b^P \simeq \frac{H \Ind^T \alpha^P}{n + \Ind^T \alpha^P}$ and  $w^\sigma_h \simeq \frac{H}{\sqrt{n + H^2 \Ind^T\alpha^R + \Ind^T \alpha^P}}$.
Summing using a pigeonhole principle for $\sum_{s,a,h} n(s,a,h) = T$ gives us an upper bound on the regret.
We write
$K(s,a,h) := \left(\alpha_1^R(s,a,h) + H \Ind^T \alpha^P(s,a,h) \right)$
} to bound the effects of the prior mistmatch in RLSVI arising from the bias terms $b^R, b^P$.
The constraint $\alpha^T\Ind \ge 2$ can only be violated twice for each $s,a,h$.
Therefore up to $O(\cdot)$ notation:
\vspace{-1mm}
{\small
\medmuskip=0mu
\thinmuskip=0mu
\thickmuskip=0mu
\begin{eqnarray*}
    & \hspace{-13mm} \Exp \left[\sum_{l=0}^{T/H-1} \Exp[ \Delta^{\rm conc}_l | \Hc_l ] \right]
        \le  \ \ 2SAH + \\
    & \sum_{s,a,h}  K(s,a,h) \log(T + K(s,a,h)) + H \sqrt{SAHT \log(T)}
\end{eqnarray*}
}
\vspace{-4mm}
This completes the proof of Theorem \ref{thm: regret}.

\section{Experiments}

Our analysis in Section \ref{sec: proof tabular} shows that RLSVI with tabular basis functions acts as an effective Gaussian approximation to PSRL.
This demonstrates a clear distinction between exploration via randomized value functions and dithering strategies such as Example \ref{example:example1}.
However, the motivating for RLSVI is not for tabular environments, where several provably efficient RL algorithms already exist, but instead for large systems that require generalization.

We believe that, under some conditions, it may be possible to establish polynomial regret bounds for RLSVI with value function generalization.
To stimulate thinking on this topic we present a conjecture of result what may be possible in Appendix \ref{app: conjecture}.
For now, we will present a series of experiments designed to test the applicability and scalability of RLSVI for exploration with generalization.

Our experiments are divided into three sections.
First, we present a series of didactic chain environments similar to Figure \ref{fig:example1}.
We show that RLSVI can effectively synthesize exploration with generalization with both coherent and agnostic value functions that are intractable under any dithering scheme.
Next, we apply our Algorithm to learning to play Tetris.
We demonstrate that RLSVI leads to faster learning, improved stability and a superior learned policy in a large-scale video game.
Finally, we consider a business application with a simple model for a recommendation system.
We show that an RL algorithm can improve upon even the optimal myopic bandit strategy.
RLSVI learns this optimal strategy when dithering strategies do not.

\vspace{-2mm}
\subsection{Testing for efficient exploration}
\label{sec: chain}

We now consider a series of environments modelled on Example \ref{example:example1}, where dithering strategies for exploration are provably inefficient.
Importantly, and unlike the tabular setting of Section \ref{sec: proof tabular}, our algorithm will only interact with the MDP but through a set of basis function $\Phi$ which generalize across states.
We examine the empirical performance of RLSVI and find that it does efficiently balance exploration and generalization in this didactic example.

\subsubsection{Coherent learning}

{\medmuskip=1mu
\thinmuskip=1mu
\thickmuskip=1mu
In our first experiments, we generate a random set of $K$ basis functions.
This basis is coherent but the individual basis functions are not otherwise informative.
We form a random linear subspace $V_{hK}$ spanned by $(\Ind, Q^*_h, \tilde{w}_1, .., \tilde{w}_{k-2})$.
Here $w_i$ and $\tilde{w}_i$ are IID Gaussian $\sim N(0,I) \in \Real^{SA}$.
We then form $\Phi_h$ by projecting $(\Ind, w_1, .., w_{k-1})$ onto $V_{hK}$ and renormalize each component to have equal 2-norm\footnote{For more details on this experiment see Appendix \ref{app: chain experiments}.}.
Figure \ref{fig: chain performance} presents the empirical regret for RLSVI with $K=10, N=50, \sigma=0.1, \lambda=1$ and an $\epsilon$-greedy agent over 5 seeds\footnote{In this setting any choice of $\epsilon$ or Boltzmann $\eta$ is equivalent.}.
}

\begin{figure}[h!]
\vspace{-4mm}
\centering
    \subfigure[First $2000$ episodes]
        {\includegraphics[width=0.4\linewidth]{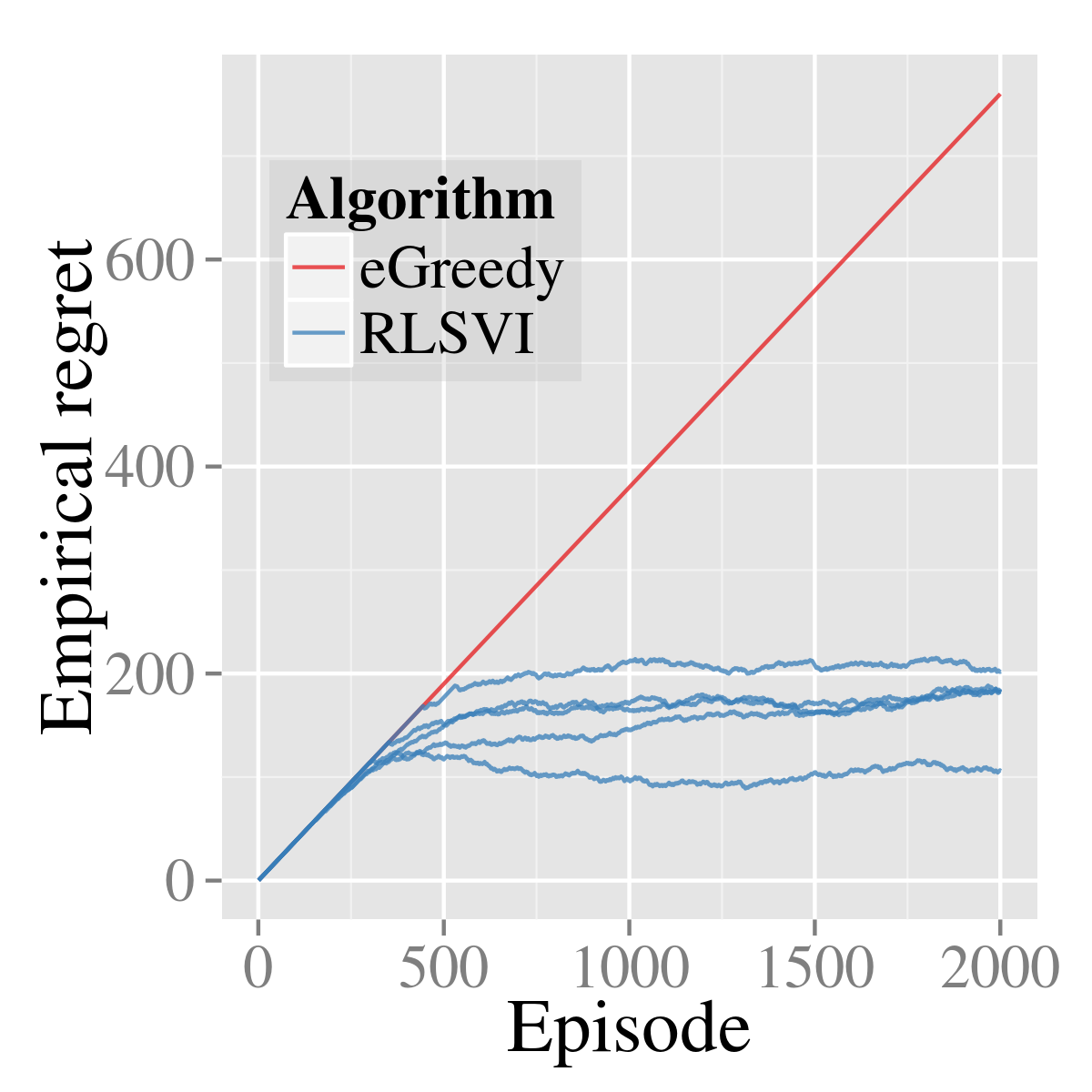}
        \label{fig:short regret}}
    \hspace{0.05\linewidth}
    \subfigure[First $10^6$ episodes]
        {\includegraphics[width=0.4\linewidth]{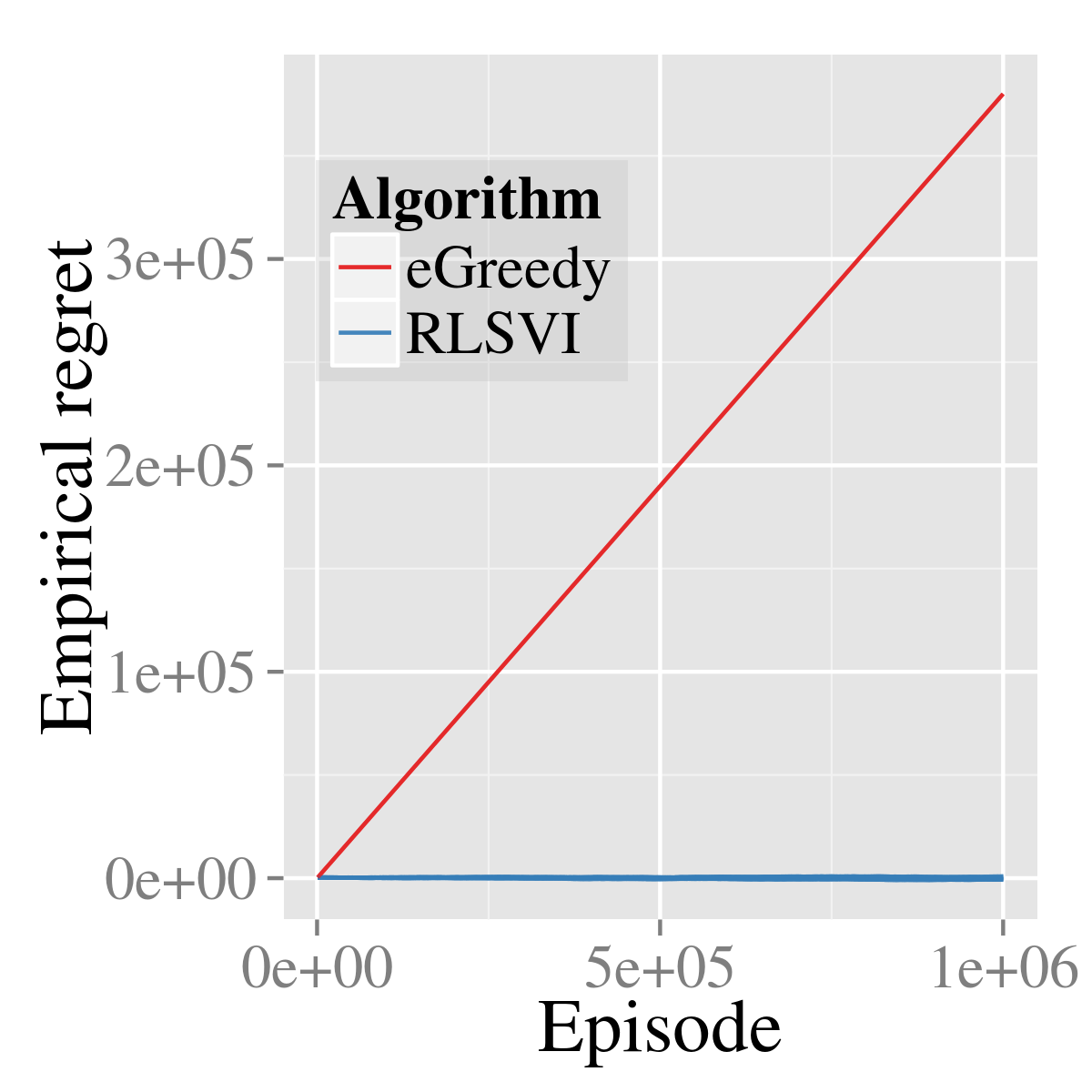}
        \label{fig:long regret}}
    \vspace{-4mm}
    \caption{Efficient exploration on a 50-chain}
    \label{fig: chain performance}
\vspace{-2mm}
\end{figure}

Figure \ref{fig:example1} shows that RLSVI consistently learns the optimal policy in roughly $500$ episodes.
Any dithering strategy would take at least $10^{15}$ episodes for this result.
The state of the art upper bounds for the efficient optimistic algorithm UCRL given by appendix C.5 in \cite{dann2015sample}  for
{\medmuskip=2mu
\thinmuskip=1mu
\thickmuskip=3mu
$H=15, S=6, A=2, \epsilon=1, \delta=1$}
only kick in after more than $10^{10}$ suboptimal episodes.
RLSVI is able to effectively exploit the generalization and prior structure from the basis functions to learn much faster.

We now examine how learning scales as we change the chain length $N$ and number of basis functions $K$.
We observe that RLSVI essentially maintains the optimal policy once it discovers the rewarding state.
We use the number of episodes until 10 rewards as a proxy for learning time.
We report the average of five random seeds.

\begin{figure}[!h]
\vspace{-2mm}
\centering
\includegraphics[width=0.9\linewidth]{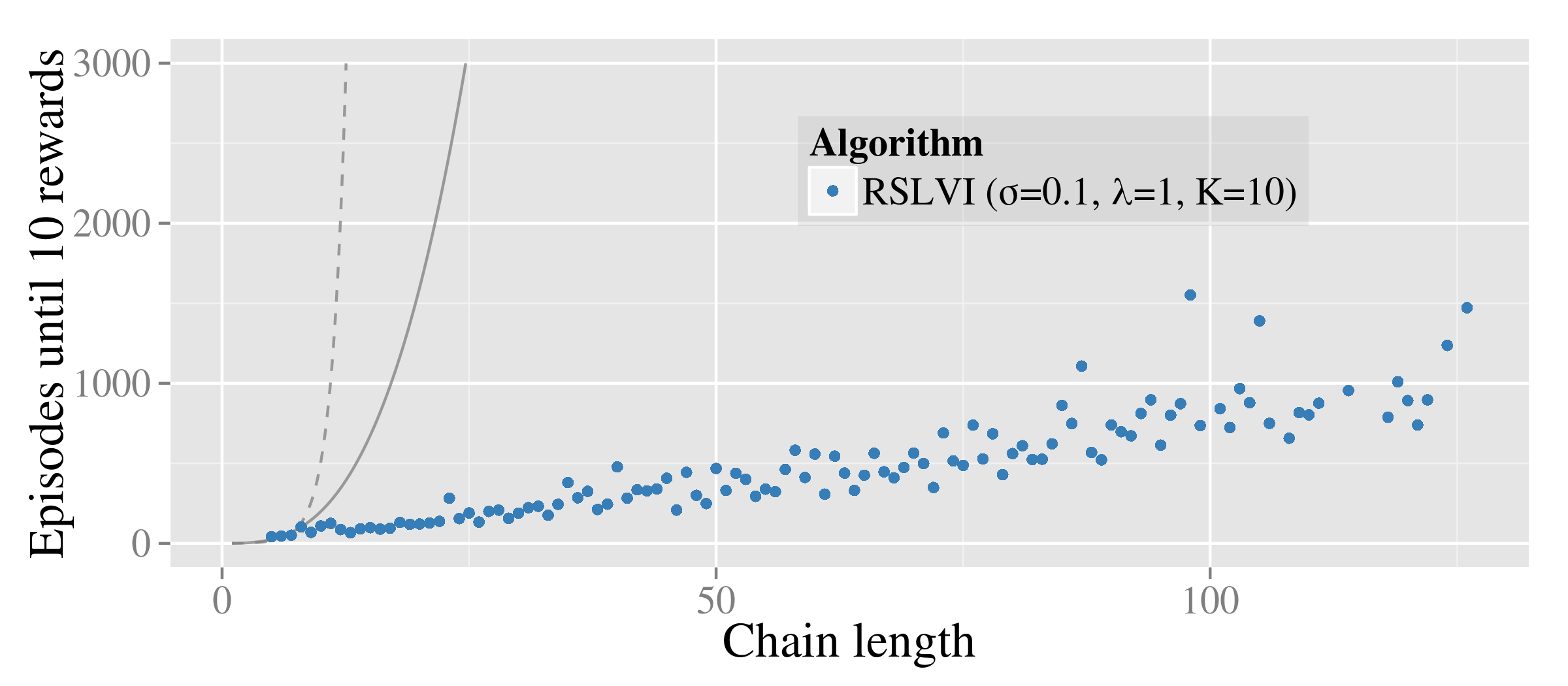}
\vspace{-5mm}
\caption{\small RLSVI learning time against chain length.}
\label{fig: small chainLen}
\vspace{-2mm}
\end{figure}

{\medmuskip=0mu
\thinmuskip=0mu
\thickmuskip=0mu
Figure \ref{fig: small chainLen} examines the time to learn as we vary the chain length $N$ with fixed $K=10$ basis functions.
We include the dithering lower bound $2^{N-1}$ as a dashed line and a lower bound scaling $\frac{1}{10} H^2SA$ for tabular learning algorithms as a solid line \cite{dann2015sample}.
For $N=100$, $2^{N-1} > 10^{28}$ and $H^2SA > 10^6$.
RLSVI demonstrates scalable generalization and exploration to outperform these bounds.
}

\begin{figure}[!h]
\vspace{-2mm}
\centering
\includegraphics[width=0.9\linewidth]{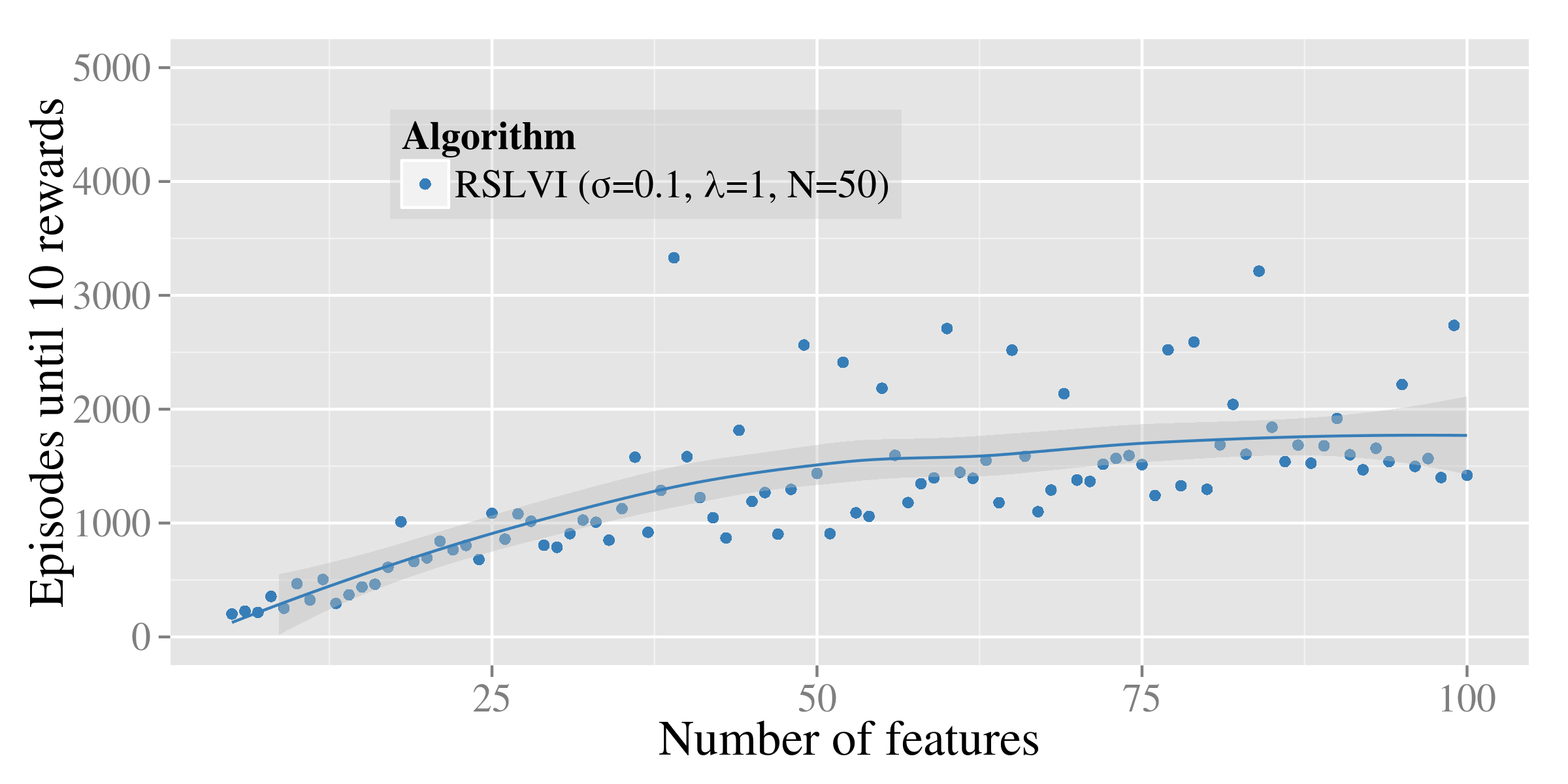}
\vspace{-4mm}
\caption{\small RLSVI learning time against number of basis features.}
\label{fig: small nFeat}
\vspace{-2mm}
\end{figure}

{\medmuskip=0mu
\thinmuskip=0mu
\thickmuskip=0mu
Figure \ref{fig: small nFeat} examines the time to learn as we vary the basis functions $K$ in a fixed $N=50$ length chain.
Learning time scales gracefully with $K$.
Further, the marginal effect of $K$ decrease as ${\rm dim}(V_{hK})=K$ approaches ${\rm dim}(\Real^{SA}) = 100$.
We include a local polynomial regression in blue to highlight this trend.
Importantly, even for large $K$ the performance is far superior to the dithering and tabular bounds\footnote{For chain $N=50$, the bounds $2^{N-1} > 10^{14}$ and $H^2SA > 10^5$.}.
}

\begin{figure}[!h]
\vspace{-2mm}
\centering
\includegraphics[width=0.9\linewidth]{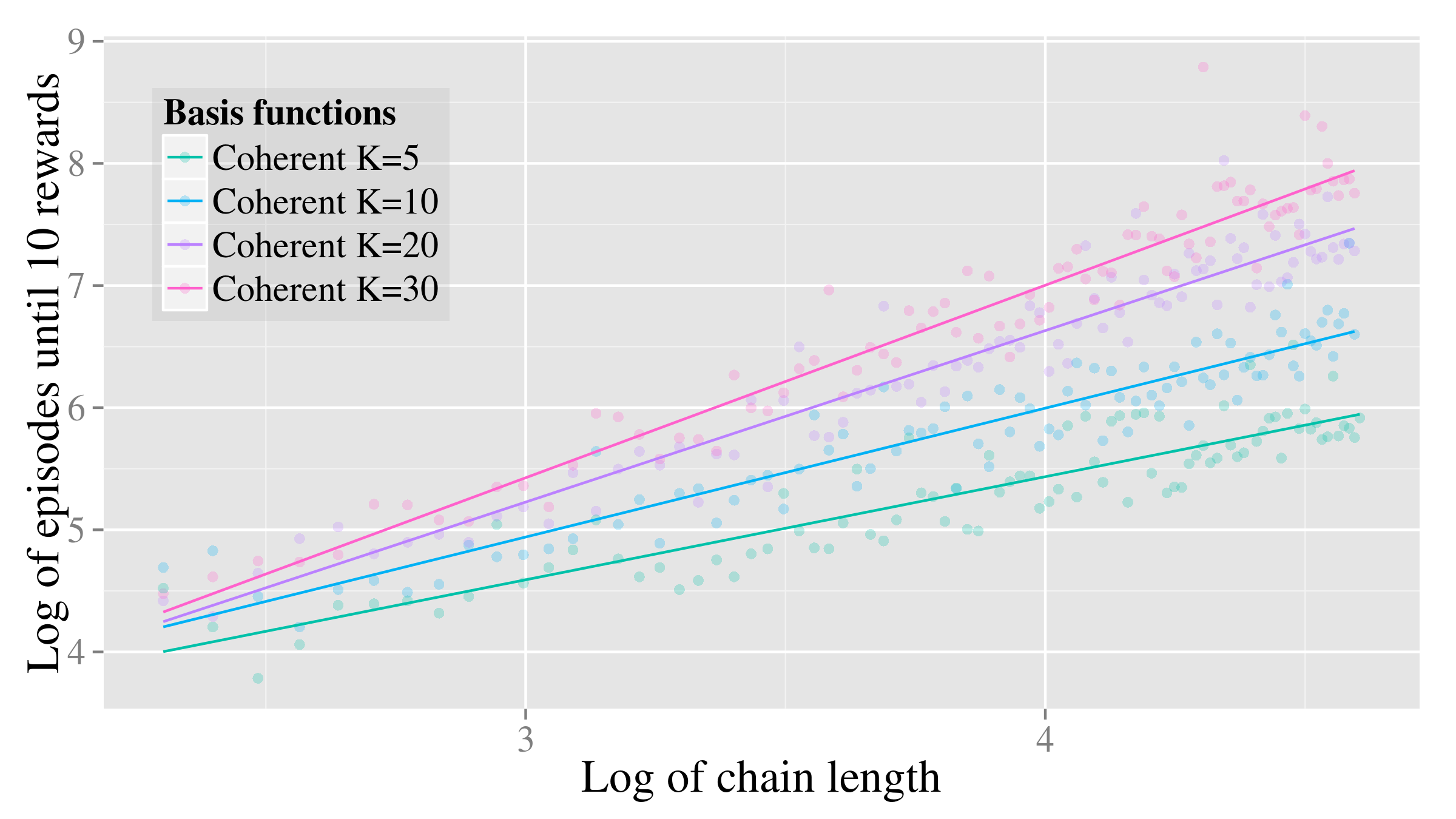}
\vspace{-3mm}
\caption{\small Empirical support for polynomial learning in RLSVI.}
\label{fig: log chain}
\vspace{-2mm}
\end{figure}

Figure \ref{fig: log chain} examines these same scalings on a logarithmic scale.
We find the data for these experiments is consistent with polynomial learning as hypothesized in Appendix \ref{app: conjecture}.
These results are remarkably robust over several orders of magnitude in both $\sigma$ and $\lambda$.
We present more detailed analysis of these sensitivies in Appendix \ref{app: chain experiments}.

\subsubsection{Agnostic learning}

Unlike the example above, practical RL problems will typically be agnostic.
The true value function $Q^*_h$ will not lie within $V_{hK}$.
To examine RLSVI in this setting we generate basis functions by adding Gaussian noise to the true value function $\phi_{hk} \sim N(Q^*_h, \rho I)$.
The parameter $\rho$ determines the scale of this noise.
For $\rho=0$ this problem is coherent but for $\rho > 0$ this will typically not be the case.
We fix $N=20, K=20, \sigma=0.1$ and $\lambda=1$.

\begin{figure}[!h]
\vspace{-3mm}
\centering
\includegraphics[width=0.95\linewidth]{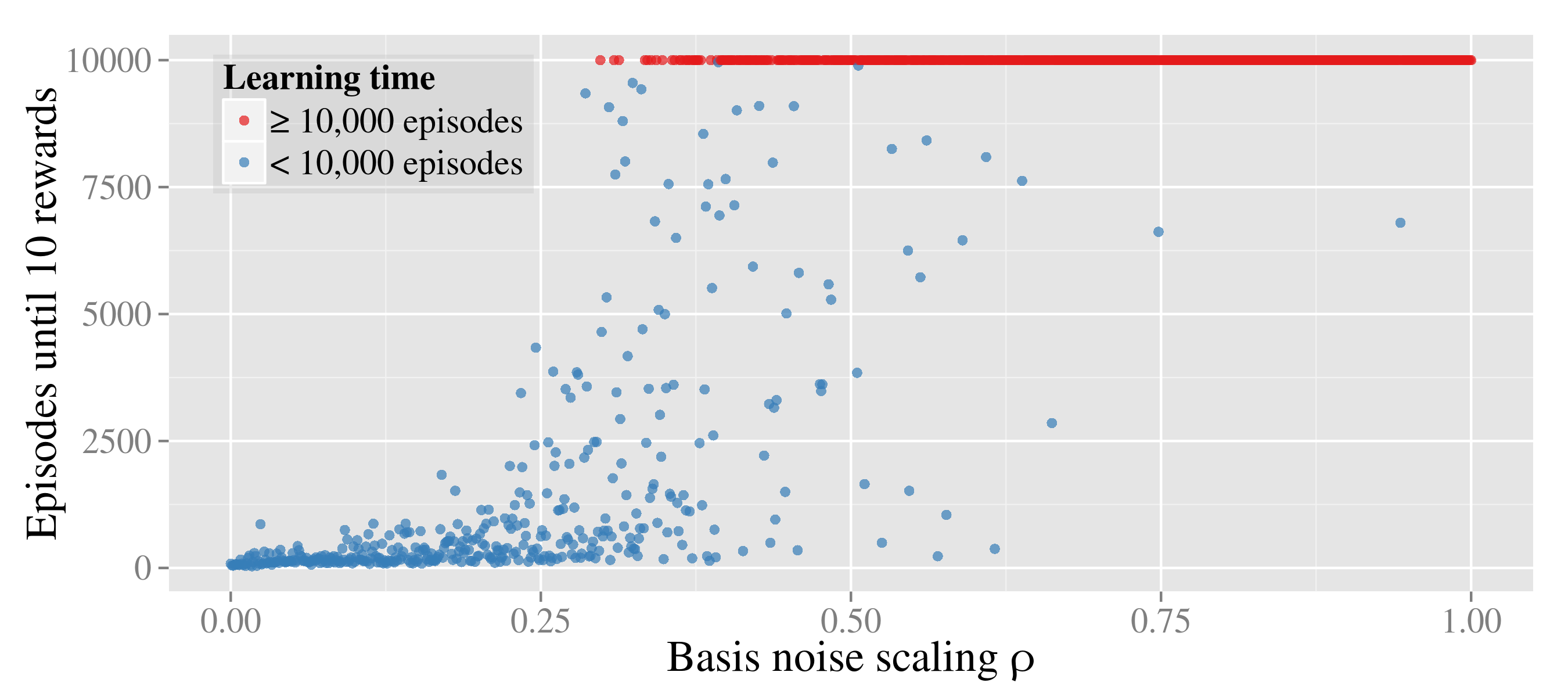}
\vspace{-4mm}
\caption{\small RLSVI is somewhat robust model mispecification.}
\label{fig: rho sweep}
\vspace{-2mm}
\end{figure}

For {\medmuskip=0mu
\thinmuskip=0mu
\thickmuskip=0mu$i=0,..,1000$ we run RLSVI for 10,000 episodes with $\rho = i/1000$ and a random seed.
Figure \ref{fig: rho sweep} presents the number of episodes until 10 rewards for each value of $\rho$.
For large values of $\rho$}, and an extremely misspecified basis, RLSVI is not effective.
However, there is some region $0 < \rho < \rho^*$ where learning remains remarkably stable\footnote{\medmuskip=0mu
\thinmuskip=0mu
\thickmuskip=0mu
Note $Q^*_h(s,a) \in \{0,1\}$ so $\rho=0.5$ represents significant noise.}.

This simple example gives us some hope that RLSVI can be useful in the agnostic setting.
In our remaining experiments we will demonstrate that RLSVI can acheive state of the art results in more practical problems with agnostic features.

\vspace{-2mm}
\subsection{Tetris}
\label{sec: tetris}

We now turn our attention to learning to play the iconic video game Tetris.
In this game, random blocks fall sequentially on a 2D grid with 20 rows and 10 columns.
At each step the agent can move and rotate the object subject to the constraints of the grid.
The game starts with an empty grid and ends when a square in the top row becomes full.
However, when a row becomes full it is removed and all bricks above it move downward.
The objective is to maximize the score attained (total number of rows removed) before the end of the game.

Tetris has been something of a benchmark problem for RL and approximate dynamic programming, with several papers on this topic \cite{gabillon2013approximate}.
Our focus is not so much to learn a high-scoring Tetris player, but instead to demonstrate the RLSVI offers benefits over other forms of exploration with LSVI.
Tetris is challenging for RL with a huge state space with more than $2^{200}$ states.
In order to tackle this problem efficiently we use 22 benchmark features.
These featurs give the height of each column, the absolute difference in height of each column, the maximum height of a column, the number of ``holes'' and a constant.
It is well known that you can find far superior linear basis functions, but we use these to mirror their approach.

{
\medmuskip=0mu
\thinmuskip=0mu
\thickmuskip=0mu
In order to apply RLSVI to Tetris, which does not have fixed episode length, we made a few natural modifications to the algorithm.
First, we approximate a time-homogeneous value function.
We also only the keep most recent $N=10^5$ transitions to cap the linear growth in memory and computational requirements, similar to \cite{mnih2015human}.
Details are provided in Appendix \ref{app: tetris experiments}.
In Figure \ref{fig: tetris 20} we present learning curves for RLSVI $\lambda=1, \sigma=1$ and LSVI with a tuned $\epsilon$-greedy exploration schedule\footnote{We found that we could not acheive good performance for any fixed $\epsilon$. We used an annealing exploration schedule that was tuned to give good performance. See Appendix \ref{app: tetris experiments}} averaged over 5 seeds.
The results are significant in several ways.
}

First, both RLSVI and LSVI make significant improvements over the previous approach of LSPI with the same basis functions \cite{bertsekas1996temporal}.
Both algorithms reach higher final performance ($\simeq 3500$ and $4500$ respectively) than the best level for LSPI ($3183$).
They also reach this performance after many fewer games and, unlike LSPI do not ``collapse'' after finding their peak performance.
We believe that these improvements are mostly due to the memory replay buffer, which stores a bank of recent past transitions, rather than LSPI which is purely online.

Second, both RLSVI and LSVI learn from scratch where LSPI required a scoring initial policy to begin learning.
We believe this is due to improved exploration schemes, LSPI is completely greedy so struggles to learn without an initial policy.
LSVI with a tuned $\epsilon$ schedule is much better.
However, we do see a significant improvement through exploration via RLSVI even when compared to the tuned $\epsilon$ scheme.
More details are available in Appendix \ref{app: tetris experiments}.

\begin{figure}[!h]
\vspace{-2mm}
\centering
\includegraphics[width=0.9\linewidth]{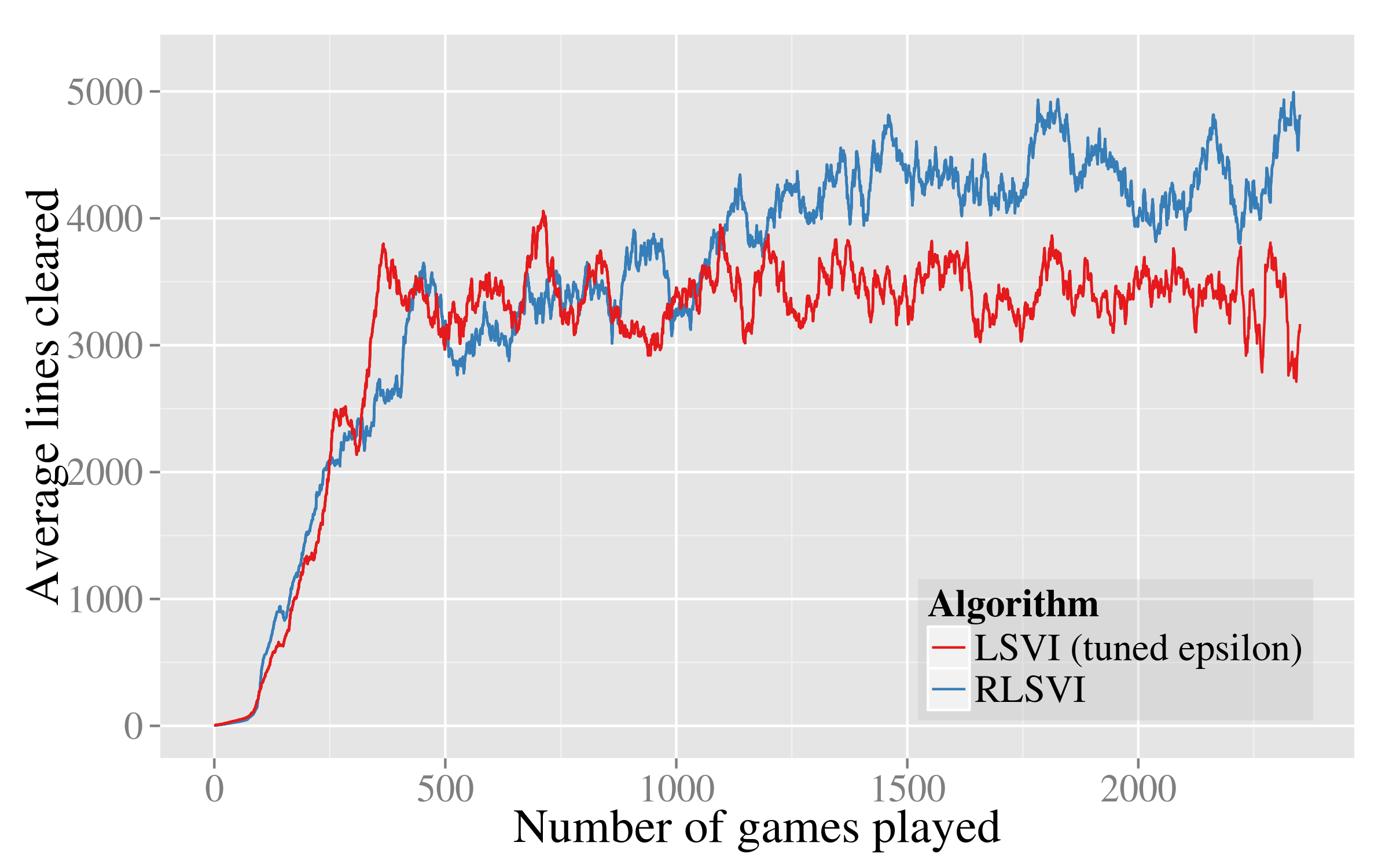}
\vspace{-4mm}
\caption{\small Learning to play Tetris with linear Bertsekas features.}
\vspace{-3mm}
\label{fig: tetris 20}
\end{figure}

\subsection{A recommendation engine}
\label{sec: recommendation}

We will now show that efficient exploration and generalization can be helpful in a simple model of customer interaction.
Consider an agent which recommends $J \leq N$ products from $\mathcal{Z} =\left \{ 1,2, \ldots, N \right \}$ sequentially to a customer.
The conditional probability that the customer likes a product depends on the product, some items are better than others.
However it also depends on what the user has observed, what she liked and what she disliked.
We represent the products the customer has seen by $\tilde{\Zc} \subseteq \Zc$.
For each product $n \in \tilde{\Zc}$ we will indicate $x_n \in \{-1, +1\}$ for her preferences $\{\text{dislike, like}\}$ respectively.
If the customer has not observed the product $n \notin \tilde{\Zc}$ we will write $x_n = 0$.
We model the probability that the customer will like a new product $a \notin \tilde{\Zc}$ by a logistic transformation linear in $x$:
\begin{equation}
\label{eqn:recommendation}
\mathbb{P} (a | x) = 1/
\left( 1+ \exp \left( - \left[ \beta_a + \textstyle \sum_{n} \gamma_{an} x_n \right] \right ) \right).
\end{equation}
Importantly, this model reflects that the customers' preferences may evolve as their experiences change.
For example, a customer may be much more likely to watch the second season of the TV show ``Breaking Bad'' if they have watched the first season and liked it.

The agent in this setting is the recommendation system, whose goal is to maximize the cumulative amount of items liked through time for each customer.
The agent does not know $p(a | x)$ initially, but can learn to estimate the parameters $\beta, \gamma$ through interactions across different customers.
Each customer is modeled as an episode with horizon length $H=J$ with a ``cold start'' and no previous observed products $\tilde{\Zc} = \emptyset$.
For our simulations we set $\beta_a=0$ $\forall a$ and sample a random problem instance by sampling $\gamma_{an} \sim N(0, c^2)$ independently for each $a$ and $n$.

\begin{figure}[!h]
\centering
\includegraphics[width=0.95\linewidth]{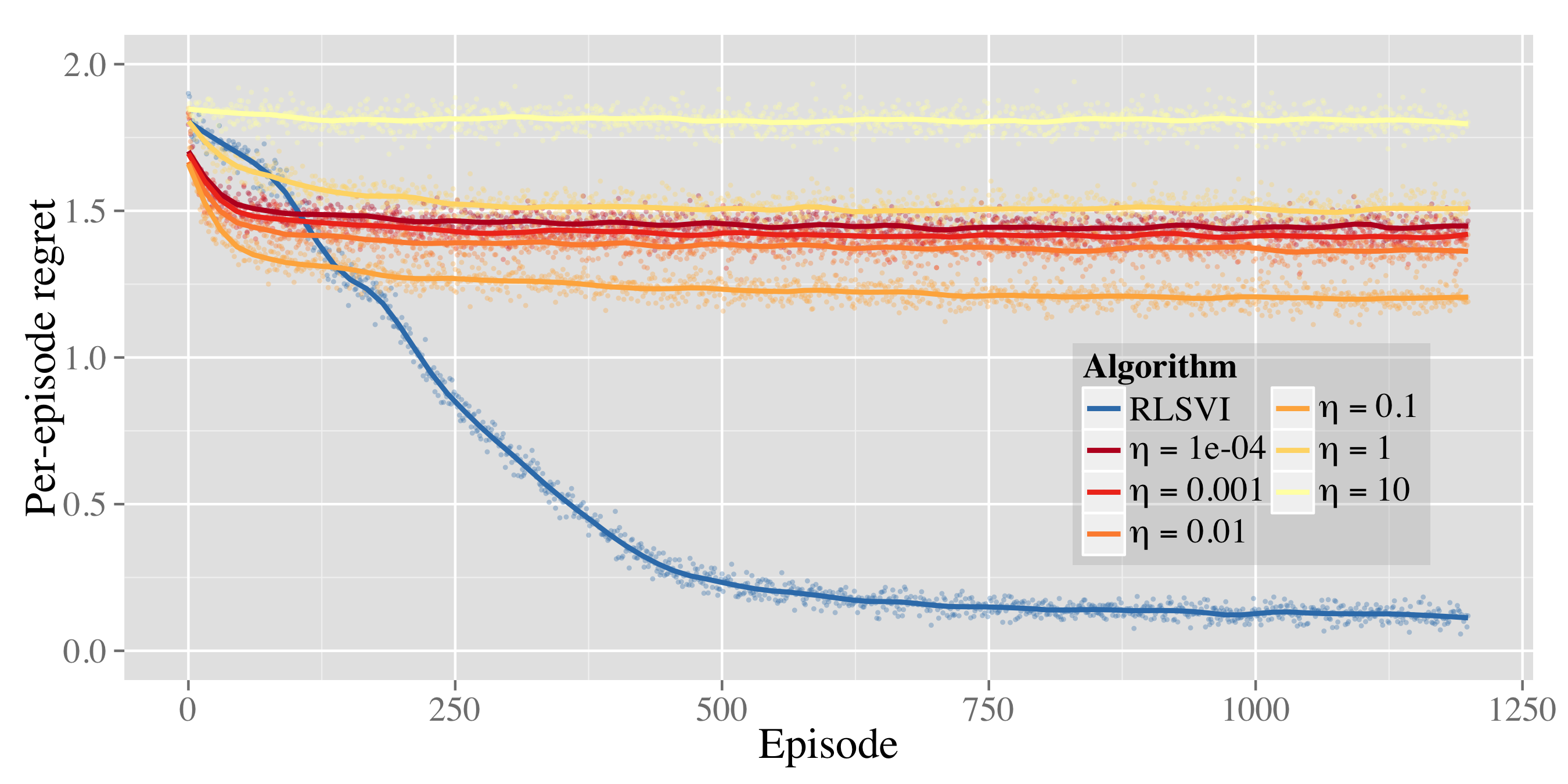}
\vspace{-4mm}
\caption{\small RLSVI performs better than Boltzmann exploration.}
\label{fig:recsys1}
\end{figure}
\vspace{-1mm}
\begin{figure}[!h]
\vspace{-1mm}
\centering
\includegraphics[width=0.95\linewidth]{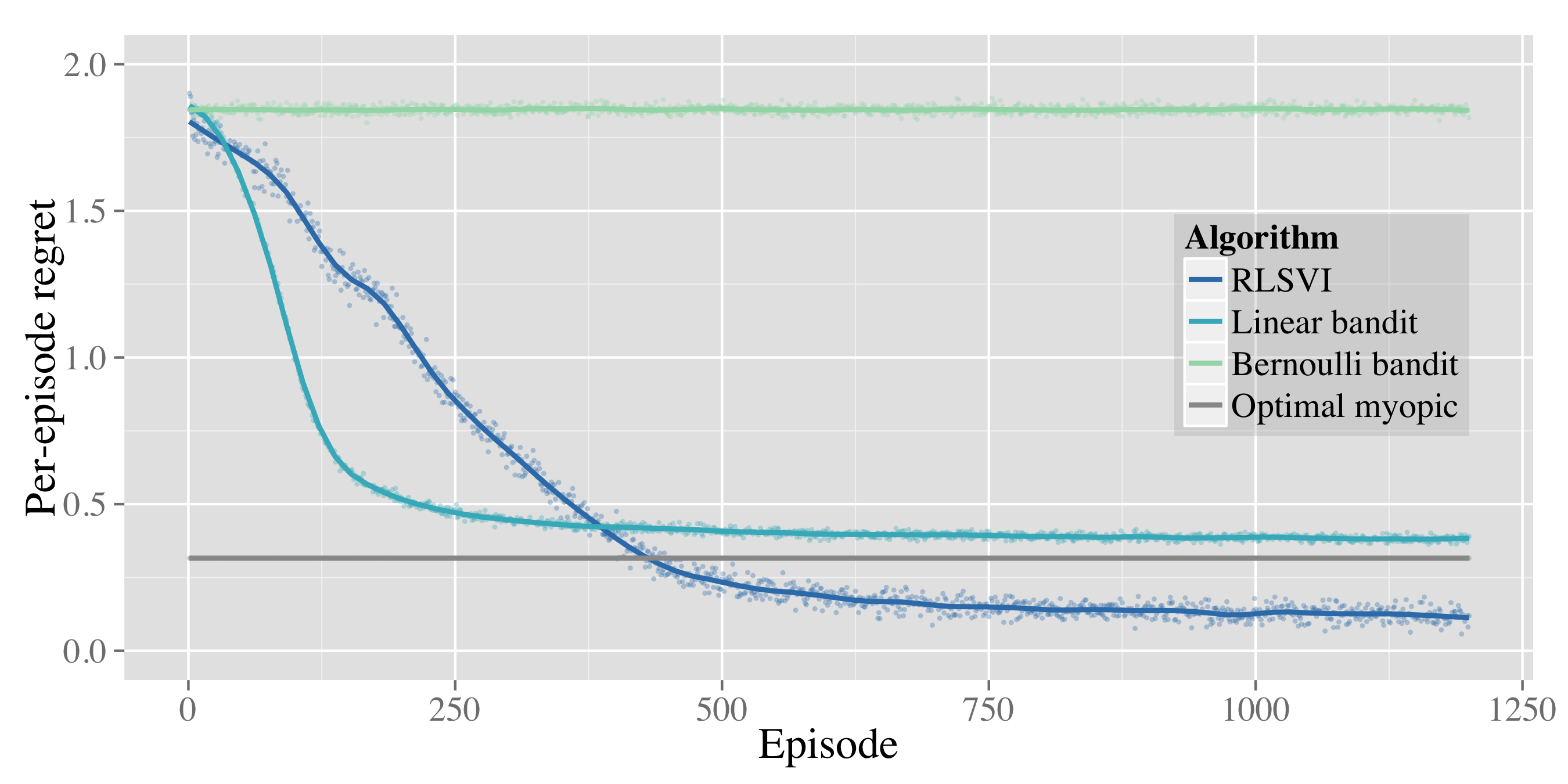}
\vspace{-4mm}
\caption{\small RLSVI can outperform the optimal myopic policy.}
\label{fig:recsys2}
\end{figure}

Although this setting is simple, the number of possible states $|\state| = |\{-1, 0, +1\}|^H = 3^J$ is exponential in $J$.
To learn in time less than $|\Sc|$ it is crucial that we can exploit generalization between states as per equation \eqref{eqn:recommendation}.
For this problem we constuct the following simple basis functions: $\forall 1 \leq n,m,a \leq N$, let $\phi_m (x,a)=\mathbf{1}\{a=m\}$ and
$\phi_{mn}(x,a)=x_n \mathbf{1}\{a=m\}$.
In each period $h$ form $\Phi_h = \left( (\phi_n)_n, (\phi_m)_m \right)$.
The dimension of our function class $K=N^2+N$ is exponentially smaller than the number of states.
However, barring a freak event, this simple basis will lead to an agnostic learning problem.

Figure~\ref{fig:recsys1} and \ref{fig:recsys2} show the performance of RLSVI compared to several benchmark methods.
In Figure~\ref{fig:recsys1} we plot the cumulative regret of RLSVI when compared against LSVI with Boltzmann exploration and identical basis features.
We see that RLSVI explores much more efficiently than Boltzmann exploration over a wide range of temperatures.

In Figure~\ref{fig:recsys2} we show that, using this efficient exploration method, the reinforcement learning policy is able to outperform not only benchmark bandit algorithms but even the optimal myopic policy\footnote{The optimal myopic policy knows the true model defined in Equation \ref{eqn:recommendation}, but does not plan over multiple timesteps.}.
Bernoulli Thompson sampling does not learn much even after $1200$ episodes, since the algorithm does not take \emph{context} into account.
The linear contextual bandit outperforms RLSVI at first.
This is not surprising, since learning a myopic policy is simpler than a multi-period policy.
However as more data is gathered RLSVI eventually learns a richer policy which outperforms the myopic policy.

Appendix \ref{app: recsys experiments} provides pseudocode for this computational study.
We set $N=10$, $H=J=5$, $c=2$ and $L=1200$. Note that such problems have $|\state|=4521$ states;
this allows us to solve each MDP exactly so that we can compute regret.
Each result is averaged over $100$ problem instances and for each problem instance, we repeat simulations $10$ times.
The cumulative regret for both RLSVI (with $\lambda=0.2$ and $\sigma^2=10^{-3}$) and LSVI with Boltzmann exploration
(with $\lambda=0.2$ and a variety of ``temperature" settings $\eta$) are plotted in Figure \ref{fig:recsys1}.
RLSVI clearly outperforms LSVI with Boltzmann exploration.

Our simulations use an extremely simplified model.
Nevertheless, they highlight the potential value of RL over multi-armed bandit approaches in recommendation systems and other customer interactions.
An RL algorithm may outperform even even an optimal myopic system, particularly where large amounts of data are available.
In some settings, efficient generalization and exploration can be crucial.

\section{Closing remarks}
\label{sec:closing_remarks}

We have established a regret bound that affirms efficiency of RLSVI in a {\it tabula rasa learning} context.
However the real promise of RLSVI lies in its potential as an efficient method for exploration in large-scale environments with generalization.
RLSVI is simple, practical and explores efficiently in several environments where state of the art approaches are ineffective.

We believe that this approach to exploration via randomized value functions represents an important concept beyond our specific implementation of RLSVI.
RLSVI is designed for generalization with linear value functions, but many of the great successes in RL have come with highly nonlinear ``deep'' neural networks from Backgammon \cite{tesauro1995temporal} to Atari\footnote{Interestingly, recent work has been able to reproduce similar performance using linear value functions \cite{DBLP:journals/corr/LiangMTB15}.} \cite{mnih2015human}.
The insights and approach gained from RLSVI may still be useful in this nonlinear setting.
For example, we might adapt RLSVI to instead take approximate posterior samples from a nonlinear value function via a nonparametric bootstrap \cite{osband2015bootstrapped}.

\bibliography{reference}
\bibliographystyle{icml2016}

\newpage


\newpage
\hspace{4mm}
\newpage
\appendix

\begin{center}
\textbf{\Large APPENDICES}
\end{center}


\section{LSVI with Boltzmann exploration/$\epsilon$-greedy exploration}
\label{sec:Boltzmann_epsilon}
The LSVI algorithm iterates backwards over
time periods in the planning horizon, in each iteration fitting a value function to the sum of immediate rewards and value estimates of the
next period.  Each value function is fitted via least-squares: note that vectors $\theta_{lh}$ satisfy
\begin{align}
\theta_{l h} \in \argmin_{\zeta \in \Real^K} \left(\|A\zeta - b\|^2 + \lambda \|\zeta\|^2\right).
\end{align}
Notice that in Algorithm \ref{alg:lsvi}, when $l=0$, matrix $A$ and vector $b$ are empty. In this case, we simply set $\theta_{l0}= \theta_{l1}= \cdots = \theta_{l,H-1}=0$.

{
\medmuskip=1mu
\thinmuskip=0mu
\thickmuskip=2mu

\begin{algorithm}[!h]
\caption{Least-Squares Value Iteration}
\textbf{Input:} Data $\Phi(s_{i0},a_{i0}),r_{i0},..,\Phi(s_{iH-1}, a_{iH-1}), r_{iH} : i < L$\\
\textcolor{white}{.} \hspace{7.5mm} Parameter $\lambda > 0$ \\
\textbf{Output:} $\theta_{l0}, \ldots, \theta_{l,H-1}$
\begin{algorithmic}[1]
\label{alg:lsvi}
\STATE $\theta_{l H} \leftarrow  0$, $\Phi_H \leftarrow \mathbf{0}$ 
\FOR{$h=H-1, \ldots, 1, 0$}
\STATE Generate regression problem $A \in \Real^{l \times K}$, $b \in \Real^l$:
\small
\begin{equation*}
    \begin{aligned}
        & A \leftarrow \left[\begin{array}{c} \Phi_h(s_{0 h}, a_{0 h}) \\ \vdots \\ \Phi_h(s_{l-1,h}, a_{l-1,h}) \end{array}\right] \\
        & b_i \leftarrow \left\{\begin{array}{ll}
    r_{ih} + \max_\alpha  \left( \Phi_{h+1} \tilde{\theta}_{l, h+1} \right) \left( s_{i,h+1} , \alpha \right) & \text{if } h < H-1 \\
    r_{ih} + r_{i,h+1} \qquad & \text{if } h = H - 1
    \end{array}\right.
    \end{aligned}
\end{equation*}
\normalsize
\STATE Linear regression for value function
$$\theta_{l h} \leftarrow (A^\top A + \lambda I)^{-1} A^{\top} b$$
\ENDFOR
\end{algorithmic}
\end{algorithm}

RL algorithms produced by synthesizing Boltzmann exploration or $\epsilon$-greedy exploration with LSVI are presented as Algorithms \ref{alg:Boltzmann} and \ref{alg:epsilon}.
In these algorithms the ``temperature'' parameters $\eta$ in Boltzmann exploration and $\epsilon$ in $\epsilon$-greedy exploration control the degree to which random perturbations distort greedy actions.

\begin{algorithm}[!h]
\caption{LSVI with Boltzmann exploration}
\label{alg:Boltzmann}
\textbf{Input:} Features $\Phi_0,..,\Phi_{H-1}$; $\eta >0, \ \lambda >0$ \\
\vspace{-4mm}
\begin{algorithmic}[1]
\FOR{$l=0,1,\cdots$}
\STATE
Compute $\theta_{l0}, \ldots, \theta_{l,H-1}$ based on Algorithm \ref{alg:lsvi}
\STATE Observe $x_{l0}$
\FOR{$h=0,1,\ldots, H-1$}
\STATE Sample $a_{lh} \sim \E \left[ (\Phi_h \theta_{lh} ) (x_{lh}, a)/ \eta \right]$
\STATE Observe $r_{lh}$ and $x_{l,h+1}$
\ENDFOR
\ENDFOR
\end{algorithmic}
\end{algorithm}

\begin{algorithm}[!h]
\caption{LSVI with $\epsilon$-greedy exploration}
\label{alg:epsilon}
\textbf{Input:} Features $\Phi_0,..,\Phi_{H-1}$; $\epsilon>0, \ \lambda >0$ \\
\vspace{-4mm}
\begin{algorithmic}[1]
\FOR{$l=0,1,\ldots$}
\STATE
Compute $\theta_{l0}, \ldots, \theta_{l,H-1}$ using Algorithm \ref{alg:lsvi}
\STATE Observe $x_{l0}$
\FOR{$h=0,1,\cdots, H-1$}
\STATE Sample $\xi \sim \text{Bernoulli}(\epsilon)$
\IF{$\xi = 1$}
\STATE Sample $a_{lh}\sim \text{unif}(\action)$
\ELSE
\STATE Sample $a_{lh} \in \argmax_{\alpha \in \action} (\Phi_h \theta_{lh} ) (x_{lh}, \alpha)$
\ENDIF
\STATE Observe $r_{lh}$ and $x_{l,h+1}$
\ENDFOR
\ENDFOR
\end{algorithmic}
\end{algorithm}
}

\newpage

\section{Efficient exploration with generalization}
\label{app: conjecture}

Our computational results suggest that, when coupled with generalization,
RLSVI enjoys levels of efficiency far beyond what can be achieved by Boltzmann or $\epsilon$-greedy exploration.
We leave as an open problem establishing efficiency guarantees in such contexts.  To stimulate thinking on this topic, we put forth a conjecture.
\begin{conjecture}
\label{conjecture: rlsvi}
For all $\mdp=\left( \state, \action, H, P, R, \pi \right)$, $\Phi_0,\ldots,\Phi_{H-1}$, $\sigma$, and $\lambda$,
if reward distributions $R$ have support $[-\sigma,\sigma]$, there is a unique $(\theta_0,\ldots, \theta_{H-1}) \in \Real^{K \times H}$ satisfying $Q^*_h = \Phi_h \theta_h$ for $h=0,\ldots,H-1$,
and $\sum_{h=0}^{H-1} \|\theta_h\|^2 \leq \frac{K H}{\lambda}$, then there exists a polynomial ${\rm poly}$ such that
$$\reg(T, \mdp) \leq \sqrt{T} \ {\rm poly}\left(K,H,\max_{h,x,a} \|\Phi_h(x,a)\|,\sigma, 1/\lambda\right).$$
\end{conjecture}
As one would hope for from an RL algorithm that generalizes, this bound does not depend on the number of states or actions.
Instead, there is a dependence on the number of basis functions.
In Appendix \ref{app: chain experiments} we present empirical results that are consistent with this conjecture.

\section{Chain experiments}
\label{app: chain experiments}

\subsection{Generating a random coherent basis}

We present full details for Algorithm \ref{alg: coherent basis}, which generates the random coherent basis functions $\Phi_h \in \Real^{SA \times K}$ for $h=1,..,H$.
In this algorithm we use some standard notation for indexing vector elements.
For any $A \in \Real^{m \times n}$ we will write $A[i, j]$ for the element in the $i^{\rm th}$ row and $j^{\rm th}$ column.
We will use the placeholder $\cdot$ to repesent the entire axis so that, for example, $A[\cdot, 1] \in \Real^n$ is the first column of $A$.

\begin{algorithm}[!h]
\caption{Generating a random coherent basis}
\label{alg: coherent basis}
\textbf{Input:} $S, A, H, K \in \Nat$, \ \  $Q^*_h \in \Real^{SA}$ for $h=1,..,H$  \\
\textbf{Output:}  $\Phi_h \in \Real^{SA \times K}$ for $h=1,..,H$ \newline
\vspace{-3mm}
\begin{algorithmic}[1]
\STATE Sample $\Psi \sim N(0, I) \in \Real^{HSA \times K}$
\STATE Set $\Psi[\cdot, 1] \leftarrow \Ind$
\STATE Stack $Q^* \leftarrow (Q^*_1, .., Q^*_h) \in \Real^{HSA}$
\STATE Set $\Psi[\cdot, 2] \leftarrow Q^*$
\STATE Form projection $P \leftarrow \Psi (\Psi^T \Psi)^{-1} \Psi^T$
\STATE Sample $W \sim N(0, I) \in \Real^{HSA \times K}$
\STATE Set $W[\cdot, 1] \leftarrow \Ind$
\STATE Project $W_P \leftarrow P W \in \Real^{HSA \times K}$
\STATE Scale $W_P[\cdot, k] \leftarrow \frac{W_P[\cdot, k]}{\|W_P[\cdot, k]\|_2} HSA$ for $k=1,..,K$
\STATE Reshape $\Phi \leftarrow {\rm reshape}(W_P) \in \Real^{H \times SA \times K}$
\STATE Return $\Phi[h, \cdot, \cdot] \in \Real^{SA \times K}$ for $h=1,..,H$
\end{algorithmic}
\end{algorithm}
\vspace{-4mm}

The reason we rescale the value function in step (9) of Algorithm \ref{alg: coherent basis} is so that the resulting random basis functions are on a similar scale to $Q^*$.
This is a completely arbitrary choice as any scaling in $\Phi$ can be exactly replicated by similar rescalings in $\lambda$ and $\sigma$.

\vspace{-1mm}
\subsection{Robustness to $\lambda, \sigma$}
\vspace{-1mm}

\begin{figure}[!h]
\vspace{-4mm}
\centering
\includegraphics[width=0.95\linewidth]{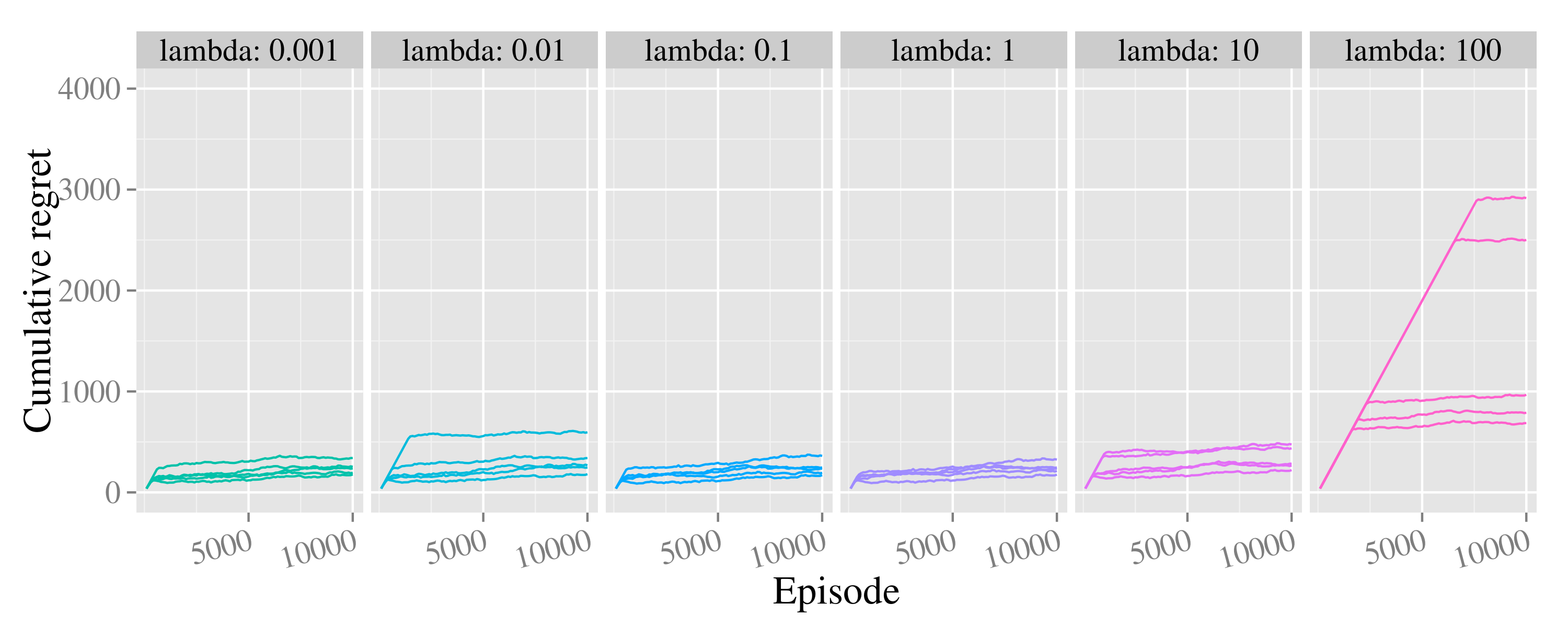}
\vspace{-4mm}
\caption{\small Fixed $\sigma=0.1$, varying $\lambda$.}
\label{fig: lambda sweep}
\vspace{-2mm}
\end{figure}

\begin{figure}[!h]
\vspace{-2mm}
\centering
\includegraphics[width=0.95\linewidth]{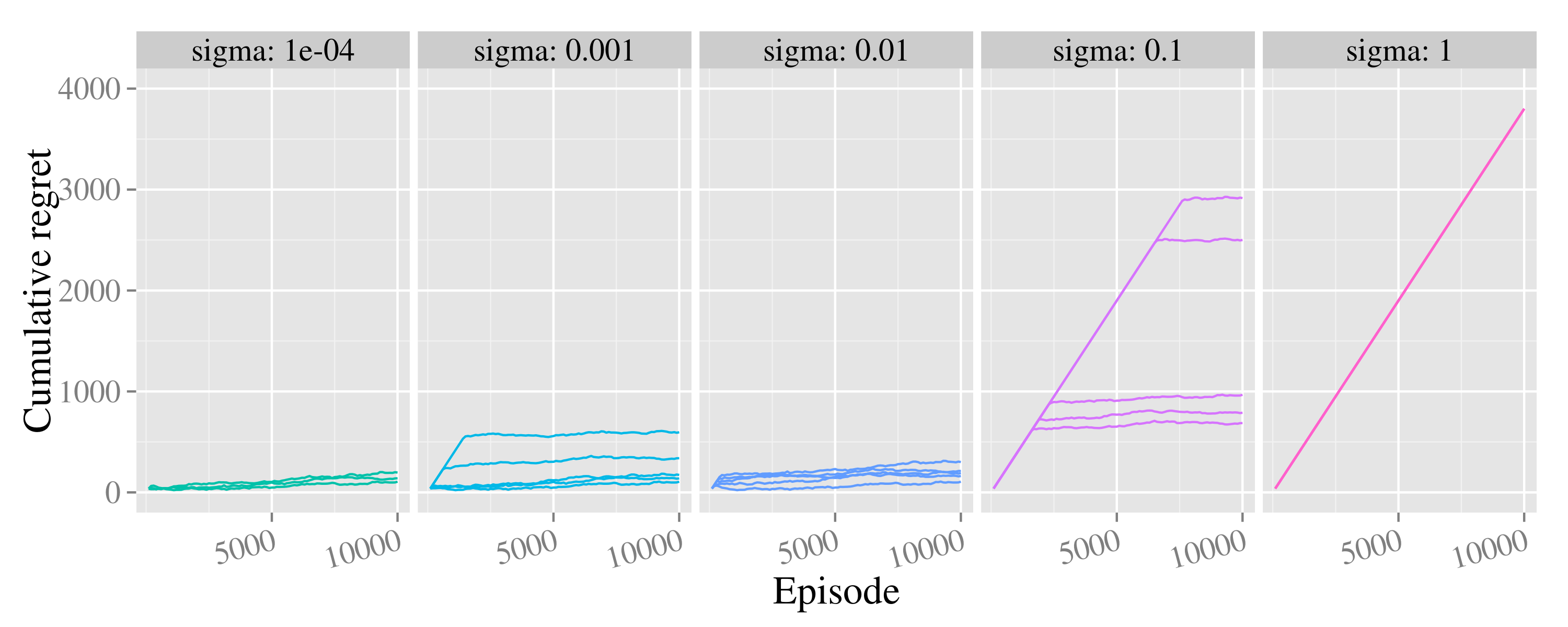}
\vspace{-4mm}
\caption{\small Fixed $\lambda=100$, varying $\sigma$.}
\label{fig: sigma sweep}
\vspace{-4mm}
\end{figure}

In Figures \ref{fig: lambda sweep} and \ref{fig: sigma sweep} we present the cumulative regret for $N=50, K=10$ over the first 10000 episodes for several orders of magnitude for $\sigma$ and $\lambda$.
For most combinations of parameters the learning remains remarkably stable.

We find that large values of $\sigma$ lead to slowers learning, since the Bayesian posterior concentrates only very slowly with new data.
However, in stochastic domains we found that choosing a $\sigma$ which is too small might cause the RLSVI posterior to concentrate too quickly and so fail to sufficiently explore.
This is a similar insight to previous analyses of Thompson sampling \cite{agrawal2012further} and matches the flavour of Theorem \ref{thm: regret}.

\vspace{-1mm}
\subsection{Scaling with number of bases $K$}
\vspace{-1mm}

In Figure \ref{fig: small nFeat} we demonstrated that RLSVI seems to scale gracefully with the number of basis features on a chain of length $N=50$.
In Figure \ref{fig: many nFeat} we reproduce these reults for chains of several different lengths.
To highlight the overall trend we present a local polynomial regression for each chain length.

\begin{figure}[!h]
\vspace{-4mm}
\centering
\includegraphics[width=0.95\linewidth]{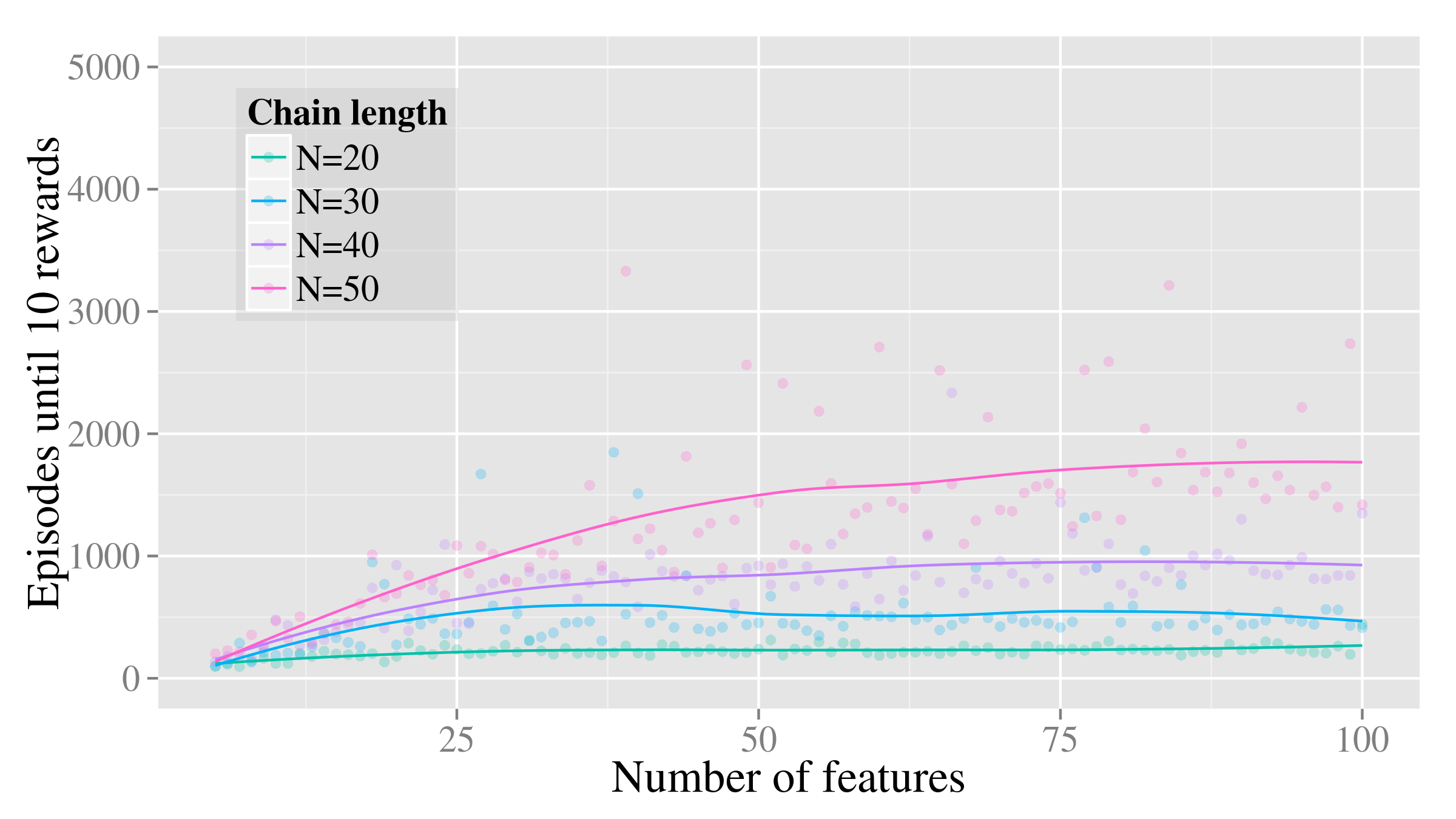}
\vspace{-4mm}
\caption{\small Graceful scaling with number of basis functions.}
\label{fig: many nFeat}
\vspace{-2mm}
\end{figure}

Roughly speaking, for low numbers of features $K$ the number of episodes required until learning appears to increase linearly with the number of basis features.
However, the marginal increase from a new basis features seems to decrease and almost plateau once the number of features reaches the maximum dimension for the problem $K \ge SA$.

\vspace{-1mm}
\subsection{Approximate polynomial learning}
\vspace{-1mm}

Our simulation results empirically demonstrate learning which appears to be polynomial in both $N$ and $K$.
Inspired by the results in Figure \ref{fig: log chain}, we present the learning times for different $N$ and $K$ together with a quadratic regression fit separately for each $K$.

\begin{figure}[!h]
\vspace{-4mm}
\centering
\includegraphics[width=0.95\linewidth]{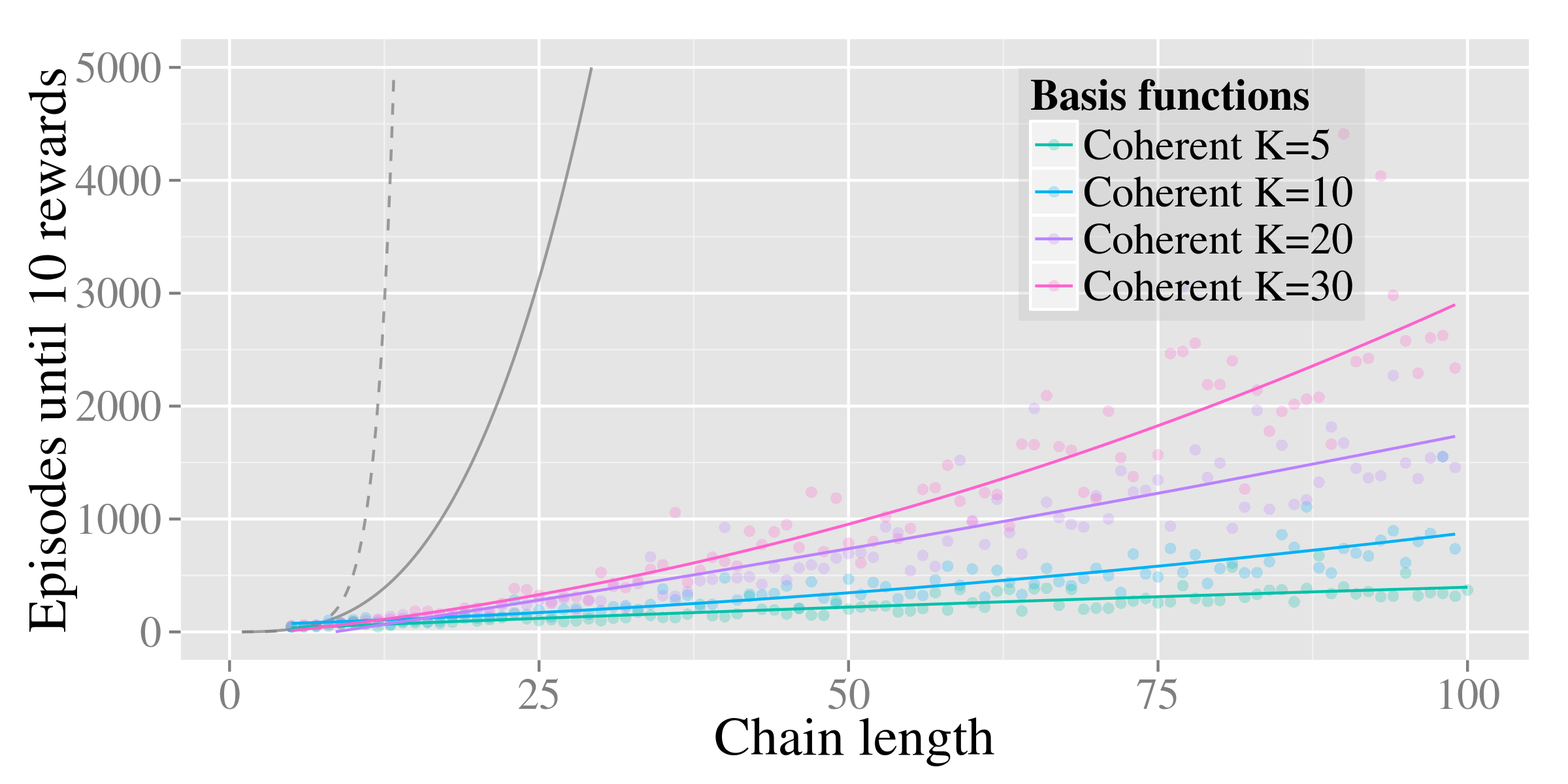}
\vspace{-4mm}
\caption{\small Graceful scaling with number of basis functions.}
\label{fig: many nFeat}
\vspace{-4mm}
\end{figure}

This is only one small set of experiments, but these results are not inconsistent with Conjecture \ref{conjecture: rlsvi}.
This quadratic model seems to fit data pretty well.

\newpage

\section{Tetris experiments}
\label{app: tetris experiments}

\subsection{Algorithm specification}

In Algorithm \ref{alg: stationary RLSVI} we present a natural adaptation to RLSVI without known episode length, but still a regular episodic structure.
This is the algorithm we use for our experiments in Tetris.
The LSVI algorithms are formed in the same way.

{
\medmuskip=1mu
\thinmuskip=0mu
\thickmuskip=2mu

\small
\begin{algorithm}[!h]
\caption{Stationary RLSVI}
\label{alg: stationary RLSVI}
\textbf{Input:} Data $\Phi(s_1,a_1),r_1, ..,\Phi(s_T, a_T)$ \\
\textcolor{white}{.} \hspace{7.5mm} Previous estimate $\tilde{\theta}_l^- \equiv \tilde{\theta}_{l-1}$ \\
\textcolor{white}{.} \hspace{7.5mm} Parameters $\lambda > 0, \  \sigma > 0, \ \gamma \in [0,1]$ \\
\textbf{Output:} $\tilde{\theta}_{l}$
\begin{algorithmic}[1]
\label{alg: stationary RLSVI}
\STATE Generate regression problem $A \in \Real^{T \times K}$, $b \in \Real^T$:
\small
\begin{equation*}
    \begin{aligned}
        & A \leftarrow \left[\begin{array}{c} \Phi_h(s_{1}, a_{1}) \\ \vdots \\ \Phi_h(s_{T}, a_{T}) \end{array}\right] \\
        & b_i \leftarrow \left\{\begin{array}{ll}
    r_{i} + \gamma \max_\alpha  \left( \Phi \tilde{\theta}_l^- \right) \left( s_{i+1} , \alpha \right) & \text{if } s_i \text{ not terminal}\\
    r_{i} \qquad & \text{if } s_i \text{ is terminal}
    \end{array}\right.
    \end{aligned}
\end{equation*}
\normalsize
\STATE Bayesian linear regression for the value function
\small
\begin{equation*}
    \begin{aligned}
        & \overline{\theta}_{l} \leftarrow \frac{1}{\sigma^2} \left(\frac{1}{\sigma^2} A^\top A + \lambda I \right)^{-1} A^{\top} b \\
        & \Sigma_{l} \leftarrow \left(\frac{1}{\sigma^2} A^\top A + \lambda I \right)^{-1}
    \end{aligned}
\end{equation*}
\normalsize
\STATE Sample $\tilde{\theta}_{l} \sim N(\overline{\theta}_{l},\Sigma_{l})$ from Gaussian posterior
\end{algorithmic}
\end{algorithm}
\normalsize
\vspace{-3mm}
\small
\begin{algorithm}[!h]
\caption{RLSVI with greedy action}
\textbf{Input:} Features $\Phi$; $\lambda >0, \ \sigma >0, \ \gamma \in [0,1]$ \\
\vspace{-4mm}
\begin{algorithmic}[1]
\label{alg:greedy}
\STATE{$\theta_0^- \leftarrow 0$; $t \leftarrow 0$}
\FOR{Episode $l=0,1,..$}
\STATE Compute $\tilde{\theta}_{l}$ using Algorithm \ref{alg: stationary RLSVI}
\STATE Observe $s_t$
\WHILE{TRUE}
\STATE Update $t \leftarrow t + 1$
\STATE Sample $a_{t} \in \argmax_{\alpha \in \action} \left( \Phi \tilde{\theta}_{} \right) \left( s_{t}, \alpha \right)$
\STATE Observe $r_{t}$ and $s_{t+1}$
\IF{$s_{t+1}$ is terminal}
\STATE BREAK
\ENDIF
\ENDWHILE
\ENDFOR
\end{algorithmic}
\end{algorithm}
\normalsize
}

This algorithm simply approximates a time-homogenous value function using Bayesian linear regression.
We found that a discount rate of $\gamma=0.99$ was helpful for stability in both RLSVI and LSVI.

In order to avoid growing computational and memory cost as LSVI collects more data we used a very simple strategy to only store the most recent $N$ transitions.
For our experiments we set $N=10^5$.
Computation for RLSVI and LSVI remained negligible compared to the cost of running the Tetris simulator for our implementations.

To see how small this memory requirement is note that, apart from the number of holes, every feature and reward is a positive integer between 0 and 20 inclusive.
The number of holes is a positive integer between 0 and 199.
We could store the information $10^5$ transitions for every possible action using less than 10mb of memory.

\subsection{Effective improvements}

We present the results for RLSVI with fixed $\sigma=1$ and $\lambda=1$.
This corresponds to a Bayesian linear regression with a known noise variance in Algorithm \ref{alg: stationary RLSVI}.
We actually found slightly better performance using a Bayesian linear regression with an inverse gamma prior over an unknown variance.
This is the conjugate prior for Gaussian regression with known variance.
Since the improvements were minor and it slightly complicates the algorithm we omit these results.
However, we believe that using a wider prior over the variance will be more robust in application, rather than picking a specific $\sigma$ and $\lambda$.

\subsection{Mini-tetris}
In Figure \ref{fig: tetris 20} we show that RLSVI outperforms LSVI even with a highly tuned annealing scheme for$\epsilon$.
However, these results are much more extreme on a didactic version of mini-tetris.
We make a tetris board with only 4 rows and only S, Z pieces.
This problem is much more difficult and highlights the need for efficient exploration in a more extreme way.

\begin{figure}[!h]
\vspace{-2mm}
\centering
\includegraphics[width=0.9\linewidth]{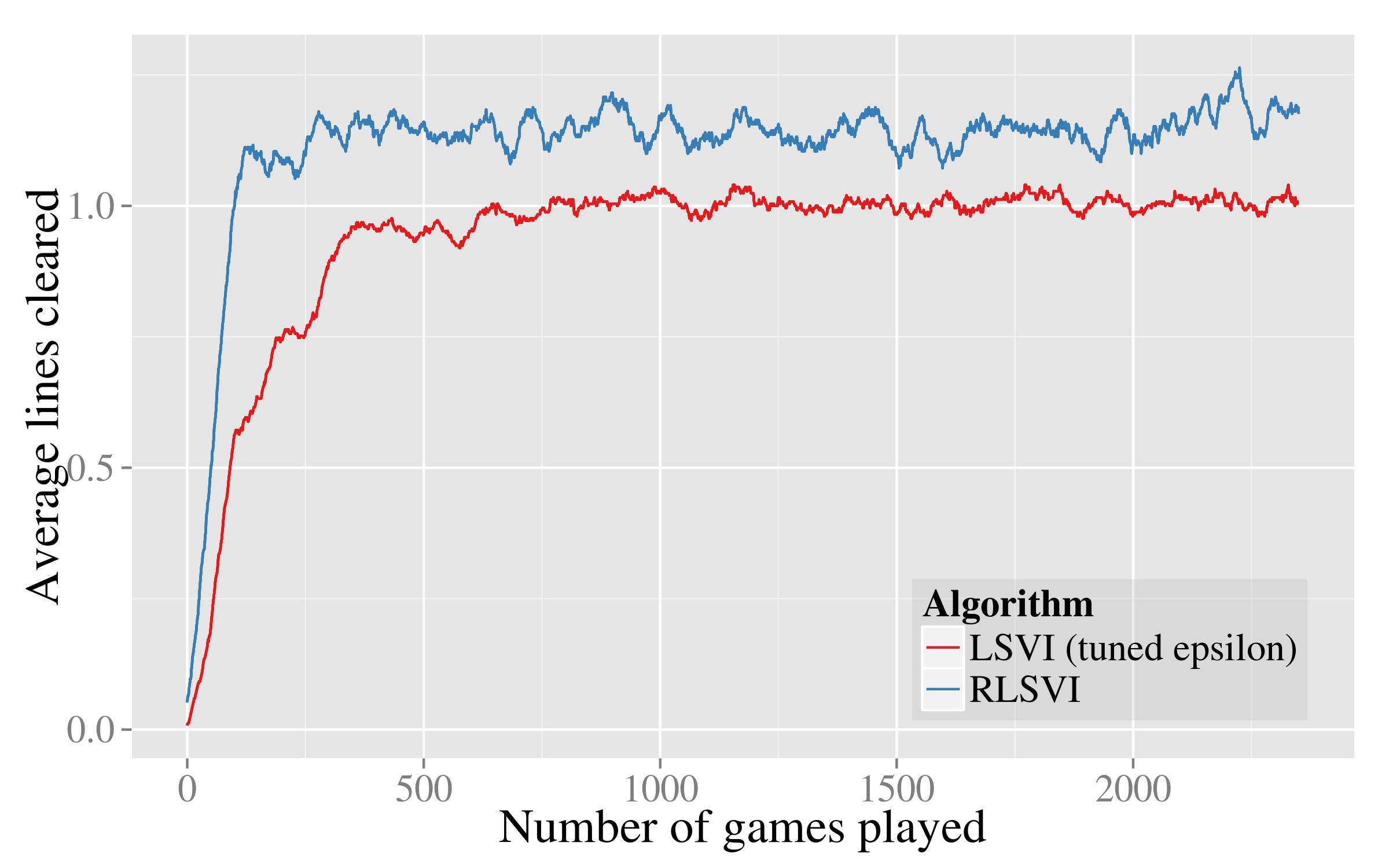}
\vspace{-3mm}
\caption{\small Reduced 4-row tetris with only S and Z pieces.}
\label{fig: tetris 5sz}
\end{figure}

In Figure \ref{fig: tetris 5sz} we present the results for this mini-tetris environment.
As expected, this example highlights the benefits of RLSVI over LSVI with dithering.
RLSVI greatly outperforms LSVI even with a tuned $\epsilon$ schedule.
RLSVI learns faster and reaches a higher convergent policy.

\newpage
\section{Recommendation system experiments}
\label{app: recsys experiments}

\subsection{Experiment Setup}
\label{app:recsys_setup}

For the recommendation system experiments, the experiment setup is specified in Algorithm \ref{alg:recsys_setup}.
We set $N=10$, $J=H=5$, $c=2$ and $L=1200$.

\begin{algorithm}[!h]
\caption{Recommendation System Experiments: Experiment Setup}
\label{alg:recsys_setup}
\textbf{Input:} $N \in \mathbb{Z}_{++}$, $J=H \in \mathbb{Z}_{++}$, $c>0$, $L \in \mathbb{Z}_{++}$ \\
\textbf{Output: $\hat{\Delta}(0), \ldots, \hat{\Delta}(L-1)$}
\begin{algorithmic}
\FOR{$i=1, \ldots, 100$}
\STATE Sample a problem instance $\gamma_{an} \sim N(0, c^2)$
\STATE Run the Bernoulli bandit algorithm $100$ times
\STATE Run the linear contextual bandit algorithm $100$ times
\FOR{for each $\eta \in \{ 10^{-4}, 10^{-3}, 10^{-2}, 10^{-1}, 1, 10\}$}
\STATE Run LSVI-Boltzmann with $\lambda=0.2$ and $\eta$ $10$ times
\ENDFOR
\STATE Run RLSVI with $\lambda=0.2$ and $\sigma^2=10^{-3}$ $10$ times
\ENDFOR
\STATE Compute the average regret for each algorithm
\end{algorithmic}
\end{algorithm}

The myopic policy is defined as follows: for all episode $l=0,1,\cdots$ and for all step $h=0,\cdots, H-1$, choose $a_{lh} \in \argmax_{a} \mathbb{P} \left(a | x_{lh} \right)$, where $a_{lh}$ and $x_{lh}$ are respectively the action and the state at step $h$ of episode $l$.

\subsection{Bernoulli bandit algorithm}
\label{app:recsys_bernoulli}
The Bernoulli bandit algorithm is described in Algorithm \ref{alg:Bernoulli}, which is a Thompson sampling algorithm with uniform prior.
Obviously, this algorithm aims to learn the myopic policy.

\begin{algorithm}[H]
\caption{Bernoulli bandit algorithm}
\label{alg:Bernoulli}
\textbf{Input:} $N \in \Nat$, $J \in \Nat$, $L \in \Nat$
\begin{algorithmic}
\STATE Initialization: Set $\alpha_n=\beta_n=1$, $\forall n=1,2,\ldots, N$
\FOR{$l=0, \ldots, L-1$}
\STATE Randomly sample $\hat{p}_{ln} \sim \mathrm{beta}\left(\alpha_n, \beta_n \right)$, $\forall n=1,\ldots,N$
\STATE Sort $\hat{p}_{ln}$'s in the descending order, and recommend the first $J$ products \textbf{in order} to the customer
\FOR{$n=1,\ldots,N$}
\IF{product $n$ is recommended in episode $l$}
\IF{customer likes product}
    \STATE $\alpha_n \leftarrow \alpha_n + 1$
\ELSE
    \STATE $\beta_n \leftarrow \beta_n + 1$
\ENDIF
\ENDIF
\ENDFOR
\ENDFOR
\end{algorithmic}
\end{algorithm}

\subsection{Linear contextual bandit algorithm}
\label{app:recsys_linear_contextual}
In this subsection, we describe the linear contextual bandit algorithm. The linear contextual bandit algorithm is similar to RLSVI, but without \emph{backward value propagation}, a key feature of RLSVI.
It is straightforward to see that the linear contextual bandit algorithm aims to learn the myopic policy. This algorithm is specified in Algorithm \ref{alg:myopic_rlsvi} and \ref{alg:LCB}. Notice that this algorithm can be implemented incrementally, hence, it is computationally efficient.
In this computational study, we use the same basis functions as RLSVI, and the same algorithm parameters (i.e. $\lambda=0.2$ and $\sigma^2=10^{-3}$).

{
\medmuskip=1mu
\thinmuskip=0mu
\thickmuskip=2mu

\small
\begin{algorithm}[!h]
\caption{Randomized exploration in linear contextual bandits}
\textbf{Input:} Data $\Phi(s_{i0},a_{i0}),r_{i0},..,\Phi(s_{iH-1}, a_{iH-1}), r_{iH} : i < L$\\
\textcolor{white}{.} \hspace{7.5mm} Parameters $\lambda > 0, \  \sigma > 0$ \\
\textbf{Output:} $\hat{\theta}_{l0}, \ldots, \hat{\theta}_{l,H-1}$
\begin{algorithmic}[1]
\label{alg:myopic_rlsvi}
\STATE $\hat{\theta}_{l H} \leftarrow  0$, $\Phi_H \leftarrow 0$
\FOR{$h=H-1, \ldots, 1, 0$}
\STATE Generate regression matrix and vector
\small
\begin{equation*}
    \begin{aligned}
        & A \leftarrow \left[\begin{array}{c} \Phi_h(s_{0 h}, a_{0 h}) \\ \vdots \\ \Phi_h(s_{l-1,h}, a_{l-1,h}) \end{array}\right] \\
        & b \leftarrow \left[\begin{array}{c} r_{0,h} \\ \vdots \\ r_{l-1,h} \end{array}\right]
    \end{aligned}
\end{equation*}
\normalsize
\STATE Estimate value function
$$\overline{\theta}_{lh} \leftarrow \frac{1}{\sigma^2} \left(\frac{1}{\sigma^2} A^\top A + \lambda \sigma^2 I \right)^{-1} A^{\top} b $$
$$\Sigma_{lh} \leftarrow \left(\frac{1}{\sigma^2 } A^\top A + \lambda I\right)^{-1}$$
\STATE Sample $\hat{\theta}_{lh} \sim N(\overline{\theta}_{lh},\Sigma_{lh})$
\ENDFOR
\end{algorithmic}
\end{algorithm}

\begin{algorithm}[!h]
\caption{Linear contextual bandit algorithm}
\textbf{Input:} Features $\Phi_0, .., \Phi_H$, $\sigma>0, \lambda>0$ \\
\vspace{-4mm}
\begin{algorithmic}[1]
\label{alg:LCB}
\FOR{$l=0,1,\cdots$}
\STATE Compute $\hat{\theta}_{l 0}, \ldots, \hat{\theta}_{l, H-1}$ using Algorithm
\ref{alg:myopic_rlsvi}
\STATE Observe $x_{l 0}$
\FOR{$h=0,\cdots, H-1$}
\STATE Sample $a_{l h} \sim \text{unif} \left( \argmax_{\alpha \in \action} \left( \Phi_h \hat{\theta}_{lh} \right) \left( x_{lh}, \alpha \right)\right)$
\STATE Observe $r_{lh}$ and $x_{l,h+1}$
\ENDFOR
\ENDFOR
\end{algorithmic}
\end{algorithm}

}

\pagebreak
\onecolumn
\section{Extensions}
\label{app: extensions}

We now briefly discuss a couple possible extensions of the version of RLSVI proposed in Algorithm \ref{alg:rlsvi} and \ref{alg:greedy}.  One is an incremental version which is computationally more efficient. The other addresses continual learning in an infinite horizon discounted Markov decision process.
In the same sense that RLSVI shares much with LSVI but is distinguished by its new approach to exploration, these extensions share much with least-squares Q-learning
\cite{Lagoudakis2002}.

\subsection{Incremental learning}
\label{sec:irlsvi}

Note that Algorithm \ref{alg:rlsvi} is a batch learning algorithm, in the sense that, in each episode $l$, though
$\Sigma_{lh}$'s can be computed incrementally, it needs all past observations to compute $\bar{\theta}_{lh}$'s.
Thus, its per-episode compute time grows with $l$, which is undesirable if the algorithm is applied over many episodes.

One way to fix this problem is to derive an incremental RLSVI that updates $\bar{\theta}_{lh}$'s and $\Sigma_{lh}$'s using summary statistics of past data and new observations made over the most recent episode.  One approach is to do this by computing
\begin{align}
\Sigma_{l+1,h}^{-1} & \leftarrow  (1- \nu_l) \Sigma_{lh}^{-1}+
\frac{1}{\sigma^2} \Phi_h \left(x_{lh}, a_{lh} \right)^{\top}  \Phi_h \left(x_{lh}, a_{lh} \right) \nonumber \\
y_{l+1,h} & \leftarrow (1-\nu_l) y_{lh} +\frac{1}{\sigma^2}
\left[ r_{lh} + \max_{\alpha \in \action} \left(\Phi_{h+1} \tilde{\theta}_{l,h+1} \right)(x_{l,h+1}, \alpha)\right]\Phi_h \left(x_{lh}, a_{lh} \right)^{\top},
\end{align}
and setting $\bar{\theta}_{l+1,h}=\Sigma_{l+1,h}^{-1} y_{l+1,h}$. Note that we sample
$\tilde{\theta}_{lh} \sim N(\bar{\theta}_{lh}, \Sigma_{lh})$, and initialize
$y_{0h}=0$, $\Sigma_{0h}^{-1}=\lambda I$, $\forall h$.   The step size
$\nu_l$ controls the influence of past observations on $\Sigma_{lh}$ and $\bar{\theta}_{lh}$.
Once $\tilde{\theta}_{lh}$'s are computed, the actions are chosen based on Algorithm \ref{alg:greedy}.
Another approach would be simply to approximate the solution for $\overline{\theta}_{lh}$ numerically via random sampling and stochastic gradient descent similar to other works with non-linear architectures \cite{mnih2015human}.
The per-episode compute time of these incremental algorithms are episode-independent, which allows for deployment at large scale.
On the other hand, we expect the batch version of RLSVI to be more data efficient and thus incur lower regret.

\subsection{Continual learning}
\label{sec:continual}
Finally,
we propose a version of RLSVI
for RL in infinite-horizon time-invariant discounted MDPs.
A discounted MDP is identified by a sextuple $\mdp=\left( \state, \action, \gamma, P, R, \pi \right)$, where $\gamma \in (0,1)$ is the discount factor.
$\state, \action, P, R, \pi$ are defined similarly with the finite horizon case. Specifically, in each time $t=0,1,\ldots$, if the state is $x_t$ and an action $a_t$ is selected then a subsequent state $x_{t+1}$ is sampled from
$P(\cdot | x_t,a_t)$ and a reward $r_t$ is sampled from $R\left(\cdot \middle | x_t,a_t,x_{t+1} \right)$. We also use $V^*$ to denote the optimal state value function, and $Q^*$ to denote the optimal action-contingent value function.
Note that $V^*$ and $Q^*$ do not depend on $t$ in this case.

\small
\begin{algorithm}[h]
\caption{Continual RLSVI}
\label{alg:rlsvi_discounted}
\textbf{Input:} $\tilde{\theta}_t \in \Real^K$, $w_t \in \Real^K$, $\Phi \in \Real^{|\state||\action| \times K}$, $\sigma >0$, $\lambda >0$, $\gamma \in (0,1)$, $\{(x_\tau, a_\tau, r_\tau) : \tau \leq t\}$, $x_{t+1}$ \\
\textbf{Output:} $\tilde{\theta}_{t+1} \in \Real^K$, $w_{t+1} \in \Real^K$
\begin{algorithmic}[1]
\STATE Generate regression matrix and vector
$$A \leftarrow \left[\begin{array}{c} \Phi(x_0, a_0) \\ \vdots \\ \Phi(x_t, a_t) \end{array}\right]
\qquad \quad
b \leftarrow \left[\begin{array}{c} r_0 + \gamma \max_{\alpha \in \action}  \left( \Phi \tilde{\theta}_t \right) \left(x_1, \alpha \right) \\ \vdots \\ r_t + \gamma \max_{\alpha \in \action}  \left( \Phi
\tilde{\theta}_t \right) \left( x_{t+1} , \alpha \right)\end{array}\right]$$
\STATE Estimate value function
$$\overline{\theta}_{t+1} \leftarrow \frac{1}{\sigma^2} \left(\frac{1}{\sigma^2} A^\top A + \lambda I\right)^{-1} A^{\top} b \qquad \quad \Sigma_{t+1} \leftarrow \left(\frac{1}{\sigma^2 } A^\top A + \lambda I\right)^{-1}$$
\STATE Sample $w_{t+1} \sim N(\sqrt{1-\gamma^2} w_t, \gamma^2 \Sigma_{t+1})$
\STATE Set $\tilde{\theta}_{t+1} = \overline{\theta}_{t+1} + w_{t+1}$
\end{algorithmic}
\end{algorithm}
\normalsize

Similarly with the episodic case,
an RL algorithm generates each action $a_{t}$ based on observations made up
to time $t$,
including all states, actions, and rewards observed in previous time steps, as well as the state space $\state$, action space $\action$, discount factor $\gamma$, and
possible prior information.
We consider a scenario in which the agent has prior knowledge that
$Q^*$ lies within a linear space spanned by a generalization matrix $\Phi \in \Real^{|\state||\action|\times K}$.

A version of RLSVI for continual learning is presented in Algorithm \ref{alg:rlsvi_discounted}.  Note that $\tilde{\theta}_t$ and $w_t$ are values computed by the algorithm in the previous time
period. We initialize $\tilde{\theta}_0=0$ and $w_0=0$.
Similarly to Algorithm \ref{alg:rlsvi},
Algorithm \ref{alg:rlsvi_discounted} randomly perturbs value estimates in
directions of significant uncertainty to incentivize exploration.
Note that the random perturbation vectors
$w_{t+1} \sim N(\sqrt{1-\gamma^2} w_t, \gamma^2 \Sigma_{t+1})$ are sampled
to ensure autocorrelation and that marginal covariance matrices of consecutive perturbations differ only slightly.
In each period $t$, once $\tilde{\theta}_t$ is computed, a greedy action is selected.  Avoiding frequent abrupt changes in the perturbation vector
is important as this allows the agent to execute on multi-period plans to reach poorly understood state-action pairs.


\section{Gaussian vs Dirichlet optimism}
\label{app: Gauss Dirichlet}

The goal of this subsection is to prove Lemma \ref{lem:Gaussian-Dirichlet}, reproduced below:

For all $v \in [0,1]^N$ and $\alpha \in [1,\infty)^N$ with $\alpha^T \Ind \ge 2$,
if $x \sim N(\alpha^\top v / \alpha^\top {\bf 1}, 1/ \alpha^\top {\bf 1})$
and $y = p^T v$ for $p \sim {\rm Dirichlet}(\alpha)$
then $x \so y$.

We begin with a lemma recapping some basic equivalences of stochastic optimism.

\begin{lemma}[Optimism equivalence]
\hspace{0.0001mm} \newline
The following are equivalent to $X \so Y$:

\begin{enumerate}[noitemsep, nolistsep]
    \item For any random variable $Z$ independent of $X$ and $Y$, $\Exp[\max(X,Z)] \ge \Exp[\max(Y,Z)]$

    \item For any $\alpha \in \Real$, $\int_\alpha^\infty \left\{ \Prob(X \ge s) - \Prob(Y \ge s) \right\} ds \ge 0$.

    \item $X =_D Y + A + W$ for $A \ge 0$ and $\Exp\left[ W | Y + A \right] = 0$ for all values $y + a$.

    \item For any $u:\Real \rightarrow \Real$ convex and increasing $\Exp[u(X)] \ge \Exp[u(Y)]$
\end{enumerate}
\end{lemma}

These properties are well known from the theory of second order stochastic dominance \cite{levy1992stochastic,hadar1969rules} but can be re-derived using only elementary integration by parts.
$X \so Y$ if and only if $-Y$ is second order stochastic dominant for $-X$.

\subsection{Beta vs. Dirichlet}

In order to prove Lemma \ref{lem:Gaussian-Dirichlet} we will first prove an intermediate result that shows a particular Beta distribution $\tilde{y}$ is optimistic for $y$.
Before we can prove this result we first state a more basic result that we will use on Gamma distributions.
\begin{lemma}
\label{le:gamma}
For independent random variables $\gamma_1 \sim \text{Gamma}(k_1, \theta)$ and $\gamma_2 \sim \text{Gamma}(k_2, \theta)$,
$$\E[\gamma_1 | \gamma_1 + \gamma_2] = \frac{k_1}{k_1 + k_2} (\gamma_1+\gamma_2)
\qquad \text{and} \qquad
\E[\gamma_2 | \gamma_1 + \gamma_2] = \frac{k_2}{k_1 + k_2} (\gamma_1+\gamma_2).$$
\end{lemma}

We can now present our optimistic lemma for Beta versus Dirichlet.
\begin{lemma}
\label{le:DirchletToBeta}
Let $y = p^\top v$ for some random variable $p \sim \text{Dirichlet}(\alpha)$ and constants $v \in \Re^d$ and $\alpha \in \mathbb{N}^d$.
Without loss of generality, assume $v_1 \leq v_2 \leq \cdots \leq v_d$.
Let $\tilde{\alpha} = \sum_{i=1}^d \alpha_i (v_i - v_1) / (v_d-v_1)$ and $\tilde{\beta} = \sum_{i=1}^d \alpha_i (v_d - v_i) / (v_d-v_1)$.
Then, there exists a random variable $\tilde{p} \sim \text{Beta}(\tilde{\alpha},\tilde{\beta})$ such that,
for $\tilde{y} = \tilde{p} v_d + (1-\tilde{p}) v_1$, $\E[\tilde{y} | y] = \E[y]$.
\end{lemma}
\proof
Let $\gamma_i = \text{Gamma}(\alpha, 1)$, with $\gamma_1,\ldots,\gamma_d$ independent, and let $\overline{\gamma} = \sum_{i=1}^d \gamma_i$, so that
$$p \equiv_D \gamma / \overline{\gamma}.$$
Let $\alpha_i^0 = \alpha_i (v_i - v_1) / (v_d-v_1)$ and $\alpha_i^1 = \alpha_i (v_d - v_i) / (v_d-v_1)$ so that
$$\alpha = \alpha^0 + \alpha^1.$$
Define independent random variables
$\gamma^0 \sim \text{Gamma}(\alpha_i^0, 1)$ and $\gamma^1 \sim \text{Gamma}(\alpha_i^1, 1)$ so that
$$\gamma \equiv_D \gamma^0 + \gamma^1.$$
Take $\gamma^0$ and $\gamma^1$ to be independent, and couple these variables with $\gamma$ so that
$\gamma = \gamma^0 + \gamma^1.$
Note that $\tilde{\beta} = \sum_{i=1}^d \alpha^0_i$ and  $\tilde{\alpha} = \sum_{i=1}^d \alpha^1_i$.
Let $\overline{\gamma}^0 = \sum_{i=1}^d \gamma^0_i$ and $\overline{\gamma}^1 = \sum_{i=1}^d \gamma^1_i$, so that
$$1-\tilde{p} \equiv_D \overline{\gamma}^0 / \overline{\gamma} \qquad \text{and} \qquad \tilde{p} \equiv_D \overline{\gamma}^1 / \overline{\gamma}.$$
Couple these variables so that
$1-\tilde{p} = \overline{\gamma}^0 / \overline{\gamma} \qquad \text{and} \qquad \tilde{p} = \overline{\gamma}^1 / \overline{\gamma}.$
We then have
{
\begin{eqnarray*}
\E[\tilde{y} | y]
&=& \E[(1- \tilde{p}) v_1 + \tilde{p} v_d | y]
= \E\left[\frac{v_1 \overline{\gamma}^0}{\overline{\gamma}} + \frac{v_d \overline{\gamma}^1}{\overline{\gamma}} \Big| y\right]
= \E\left[\E\left[\frac{v_1 \overline{\gamma}^0 + v_d \overline{\gamma}^1}{\overline{\gamma}} \Big| \gamma, y\right] \Big| y \right] \\
&=& \E\left[\frac{v_1 \E[\overline{\gamma}^0 | \gamma] + v_d \E[\overline{\gamma}^1 | \gamma]}{\overline{\gamma}} \Big| y \right]
= \E\left[\frac{v_1 \sum_{i=1}^d \E[\gamma^0_i | \gamma_i] + v_d \sum_{i=1}^d\E[\gamma^1_i | \gamma_i]}{\overline{\gamma}} \Big| y \right] \\
&\stackrel{\text{(a)}}{=}& \E\left[\frac{v_1 \sum_{i=1}^d \gamma_i \alpha_i^0 / \alpha_i + v_d \sum_{i=1}^d \gamma_i \alpha_i^1/\alpha_i}{\overline{\gamma}} \Big| y \right] \\
&=& \E\left[\frac{v_1 \sum_{i=1}^d \gamma_i (v_i - v_1) + v_d \sum_{i=1}^d \gamma_i (v_d - v_i)}{\overline{\gamma} (v_d - v_1)} \Big| y \right] \\
&=& \E\left[\frac{\sum_{i=1}^d \gamma_i  v_i}{\overline{\gamma}} \Big| y \right]
= \E\left[\sum_{i=1}^d p_i  v_i \Big| y \right]
= y,
\end{eqnarray*}
}
where (a) follows from Lemma \ref{le:gamma}.
\qed

\subsection{Gaussian vs Beta}
In the previous section we showed that a matched Beta distribution $\tilde{y}$ would be optimistic for the Dirichlet $y$.
We will now show that the Normal random variable $x$ is optimistic for $\tilde{y}$ and so complete the proof of Lemma \ref{lem:Gaussian-Dirichlet}, $x \so \tilde{y} \so y$.

Unfortunately, unlike the case of Beta vs Dirichlet it is quite difficult to show this optimism relationship between Gaussian $x$ and Beta $\tilde{y}$ directly.
Instead we make an appeal to the stronger dominance relationship of single-crossing CDFs.

\begin{definition}[Single crossing dominance]
\hspace{0.0001mm} \newline
Let $X$ and $Y$ be real-valued random variables with CDFs $F_X$ and $F_Y$ respectively.
We say that $X$ single-crossing dominates $Y$ if $\Exp[X] \ge \Exp[Y]$ and there a crossing point $a \in \Real$ such that:
\begin{equation}
    F_X(s) \ge F_Y(s) \iff s \le a.
\end{equation}
\end{definition}
Note that single crossing dominance implies stochastic optimism.
The remainder of this section is devoted to proving that the following lemma:

\begin{lemma}
\label{lem: Gauss Beta}
Let $\tilde{y} \sim Beta(\alpha, \beta)$ for any $\alpha>0, \beta>0$ and $x \sim N \left(\mu = \frac{\alpha}{\alpha + \beta}, \sigma^2 = \frac{1}{\alpha + \beta} \right)$.
Then, $x$ single crossing dominates $\tilde{y}$.
\end{lemma}
Trivially, these two distributions will always have equal means so it is enough to show that their CDFs can cross at most once on $(0,1)$.

\subsection{Double crossing PDFs}
By repeated application of the mean value theorem, if we want to prove that the CDFs cross at most once on $(0,1)$ then it is sufficient to prove that the PDFs cross at most twice on the same interval.
Our strategy will be to show via mechanical calculus that for the known densities of $x$ and $\tilde{y}$ the PDFs cross at most twice on $(0,1)$.
We lament that the proof as it stands is so laborious, but our attempts at a more elegant solution has so far been unsucessful.
The remainder of this appendix is devoted to proving this ``double-crossing'' property via manipulation of the PDFs for different values of $\alpha, \beta$.

We write $f_N$ for the density of the Normal $x$ and $f_B$ for the density of the Beta $\tilde{y}$ respectively.
We know that at the boundary $f_N(0-) > f_B(0-)$ and $f_N(1+) > f_B(1+)$ where the $\pm$ represents the left and right limits respectively.
Since the densities are postive over the interval, we can consider the log PDFs instead.
$$ l_B(x) = (\alpha - 1) \log(x) + (\beta - 1) \log(1-x) + K_B $$
$$ l_N(x) = -\frac{1}{2} (\alpha + \beta) \left(x - \frac{\alpha}{\alpha+\beta}\right)^2 + K_N$$
Since $\log(x)$ is injective and increasing, if we could show that $l_N(x) - l_B(x) = 0$ has at most two solutions on the interval we would be done.

Instead we will attempt to prove an even stronger condition, that $l'_N(x) - l'_B(x) = 0$ has at most one solution in the interval.
This is not necessary for what we actually want to show, but it is sufficient and easier to deal with since we can ignore the annoying constants.
$$ l'_B(x) = \frac{\alpha - 1}{x} - \frac{\beta - 1}{1 - x} $$
$$ l'_N(x) = \alpha - (\alpha + \beta) x $$

Finally we will consider an even stronger condition, if $l''_N(x) - l''_B(x) = 0$ has no solution then $l'_B(x) - l'_N(x)$ must be monotone over the region and so it can have at most one root.
$$ l''_B(x) = - \frac{\alpha - 1}{x^2} - \frac{\beta - 1}{(1-x)^2} $$
$$ l''_N(x) = - (\alpha + \beta) $$
So now let us define:
\begin{equation}
    h(x) := l''_N(x) - l''_B(x) = \frac{\alpha - 1}{x^2} + \frac{\beta - 1}{(1-x)^2}
        - (\alpha + \beta)
\end{equation}
Our goal now is to show that $h(x) = 0$ does not have any solutions for $x \in [0,1]$.

Once again, we will look at the derivatives and analyse them for different values of $\alpha, \beta > 0$.
$$ h'(x) = -2 \left( \frac{\alpha - 1}{x^3} - \frac{\beta - 1}{(1-x)^3} \right) $$
$$ h''(x) = 6 \left( \frac{\alpha - 1}{x^4} + \frac{\beta - 1}{(1-x)^4} \right) $$

\subsubsection{Special case $\alpha > 1$, $\beta \le 1$}

In this region we want to show that actually $g(x) = l'_N(x) - l'_B(x)$ has no solutions.
We follow a very similar line of argument and write $A = \alpha - 1 > 0$ and $B = \beta - 1 \le 0$ as before.
$$ g(x) = \alpha - (\alpha + \beta)x + \frac{\beta - 1}{1-x} - \frac{\alpha - 1}{x} $$
$$ g'(x) = h(x) = \frac{A}{x^2} + \frac{B}{(1-x)^2} - (\alpha + \beta) $$
$$ g''(x) = h'(x) = -2 \left( \frac{A}{x^3} - \frac{B}{(1-x)^3} \right) $$
Now since $B \le 0$ we note that $g''(x) \le 0$ and so $g(x)$ is a concave function.
If we can show that the maximum of $g$ lies below $0$ then we know that there can be no roots.

We now attempt to solve $g'(x) = 0$:
\begin{eqnarray*}
    g'(x) &=& \frac{A}{x^2} + \frac{B}{(1-x)^2} = 0 \\
    \implies - A / B &=& \left( \frac{x}{1-x} \right)^2 \\
    \implies x &=& \frac{K}{1+K} \in (0,1)
\end{eqnarray*}
Where here we write $K = \sqrt{-A/B} > 0$.
We're ignoring the case of $B=0$ as this is even easier to show separately.
We now evaluate the function $g$ at its minimum $x_K = \frac{K}{1+K}$ and write $C = -B \ge 0$.
\begin{eqnarray*}
    g(x_K) &=& (A+1) - (A+B+2)\frac{K}{1+K} + B(1+K) - A\frac{1+K}{K} \\
    &=& - AK^2 - AK - A + BK^3 + BK^2 + BK - K^2 + K \\
    &=& - AK^2 - AK - A - CK^3 - CK^2 - CK - K^2 +K \\
    &=& - A (A/C) - A (A/C)^{1/2} - A - C (A/C)^{3/2} - C (A/C) - C(A/C)^{1/2} - A/C + (A/C)^{1/2} \\
    &=& -A^2 C^{-1} - A^{3/2}C^{-1/2} - A - A^{3/2}C^{-1/2} - A - A^{1/2}C^{1/2} - AC^{-1} + A^{1/2}C^{1/2} \\
    &=& -A^2 C^{-1} - 2 A^{3/2}C^{-1/2} - 2 A - AC^{-1} \le 0
\end{eqnarray*}
Therefore we are done with this sub proof.
The case of $\alpha \le 1, \beta > 1$ can be dealt with similarly.

\subsubsection{Convex function $\alpha > 1, \beta > 1, (\alpha-1)(\beta-1) \ge \frac{1}{9}$}
In the case of $\alpha, \beta > 1$ we know that $h(x)$ is a convex function on $(0,1)$.
So now if we solve $h'(x^*) = 0$ and $h(x^*) > 0$ then we have proved our statement.
We will write $A = \alpha - 1, B = \beta - 1$ for convenience.

We now attempt to solve $h'(x) = 0$
\begin{eqnarray*}
    h'(x) &=& \frac{A}{x^3} - \frac{B}{(1-x)^3} = 0 \\
    \implies A / B &=& \left( \frac{x}{1-x} \right)^3 \\
    \implies x &=& \frac{K}{1+K} \in (0,1)
\end{eqnarray*}
Where for convenience we have written $K = \left( A / B \right)^{1/3} > 0$.
We now evaluate the function $h$ at its minimum $x_K = \frac{K}{1+K}$.
\begin{eqnarray*}
    h(x_K) &=& A \frac{(K+1)^2}{K^2} + B(K+1)^2 - (A + B + 2) \\
    &=& A (2 / K + 1 / K^2) + B ( K^2 + 2K ) - 2 \\
    &=& 3 (A^{2/3}B^{1/3} + A^{1/3}B^{2/3}) - 2
\end{eqnarray*}
So as long as $h(x_K) > 0$ we have shown that the CDFs are single crossing.
We note a simpler characterization of $A, B$ that guarantees this condition:
$$ A,B \ge 1/3 \implies AB \ge 1/9 \implies (A^{2/3}B^{1/3} + A^{1/3}B^{2/3}) \ge 2/3 $$
And so we have shown that somehow for $\alpha, \beta$ large enough away from 1 we are OK.
Certianly we have proved the result for $\alpha, \beta \ge 4/3$.

\subsubsection{Final region $\{\alpha > 1, \ \beta > 1, \ (\alpha-1)(\beta-1) \le \frac{1}{9}\}$}
We now produce a final argument that even in this remaining region the two PDFs are at most double crossing.
The argument is really no different than before, the only difficulty is that it is not enough to only look at the derivatives of the log likelihoods, we need to use some bound on the normalizing constants to get our bounds.
By symmetry in the problem, it will suffice to consider only the case $\alpha > \beta$, the other result follows similarly.

In this region of interest, we know that $\beta \in (1,\frac{4}{3})$ and so we will make use of an upper bound to the normalizing constant of the Beta distribution, the Beta function.
\begin{eqnarray}
    B(\alpha, \beta) &=& \int_{x=0}^1 x^{\alpha - 1}(1-x)^{\beta-1} dx \nonumber \\
    &\le& \int_{x=0}^1 x^{\alpha-1} dx = \frac{1}{\alpha}
\end{eqnarray}
Our thinking is that, because in $\Bc$ the value of $\beta-1$ is relatively small, this approximation will not be too bad.
Therefore, we can explicitly bound the log likelihood of the Beta distribution:
$$ l_B(x) \ge \tilde{l}_B(x) := (\alpha - 1) \log(x) + (\beta - 1) \log(1-x) + \log(\alpha) $$

We will now make use of a calculus argument as in the previous sections of the proof.
We want to find two points $x_1 < x_2$ for which $h(x_i) = l''_N(x) - l''_B(x) > 0$.
Since $\alpha, \beta > 1$ we know that $h$ is convex and so for all $x \notin [x_1, x_2]$ then $h > 0$.
If we can also show that the gap of the Beta over the maximum of the normal log likelihood
\begin{equation}
    {\rm Gap: } \ l_B(x_i) - l_N(x_i) \ge f(x_i) := \tilde{l}_B(x_i) - \max_x l_N(x) > 0
\end{equation}
is positive then it must mean there are no crossings over the region $[x_1, x_2]$, since $\tilde{l}_B$ is concave and therefore totally above the maximum of $l_N$ over the whole region $[x_1, x_2]$.

Now consider the regions $x \in [0,x_1)$, we know by consideration of the tails that if there is more than one root in this segment then there must be at least three crossings.
If there are three crossings, then the second derivative of their difference $h$ must have at least one root on this region.
However we know that $h$ is convex, so if we can show that $h(x_i) > 0$ this cannot be possible.
We use a similar argument for $x \in (x_2, 1]$.
We will now complete this proof by lengthy amounts of calculus.

Let's remind ourselves of the definition:
$$ h(x) := l''_N(x) - l''_B(x) = \frac{\alpha - 1}{x^2} + \frac{\beta - 1}{(1-x)^2}
        - (\alpha + \beta) $$
For ease of notation we will write $A = \alpha - 1, B = \beta - 1$.
We note that:
$$h(x) \ge h_1(x) = \frac{A}{x^2} - (A+B+2),\ \  h(x) \ge h_2(x) = \frac{B}{(1-x)^2} - (A+B+2)$$
and we solve for $h_1(x_1) = 0, h_2(x_2)=0$.
This means that
$$x_1 = \sqrt{\frac{A}{A+B+2}}, \ \ x_2 = 1 - \sqrt{\frac{B}{A+B+2}}$$
and clearly $h(x_1) > 0, h(x_2) > 0$.
Now, if we can show that, for all possible values of $A,B$ in this region $f(x_i) = l_B(x_i) - \max_x l_N(x) > 0$, our proof will be complete.

We will now write $f(x_i) = f_i(A,B)$ to make the dependence on $A, B$ more clear.
$$ f_1(A, B) = \log(1 + A) + A \log\left( \sqrt{\frac{A}{A + B + 2}} \right) + B \log\left(1 - \sqrt{\frac{A}{A + B + 2}} \right) + \frac{1}{2} \log(2 \pi) - \frac{1}{2}\log(A + B + 2) $$
$$ f_2(A, B) = \log(1 + A) + A \log\left( 1 - \sqrt{\frac{B}{A + B + 2}} \right) + B \log\left(\sqrt{\frac{B}{A + B + 2}} \right) + \frac{1}{2} \log(2 \pi) - \frac{1}{2}\log(A + B + 2) $$

We will now show that $\frac{\partial f_i}{\partial B} \le 0$ for all of the values in our region $A > B > 0$.
\begin{eqnarray*}
\frac{\partial f_1}{\partial B} &=& -\frac{A}{2(A + B +2)}
    + \log\left(1-\sqrt{\frac{A}{A + B + 2}}\right)
    + \frac{B \sqrt{A}}{2(A+B+2)^{3/2}\left(1-\sqrt{\frac{A}{A + B + 2}}\right)}
    - \frac{1}{2(A+B+2)} \\
    &=& \frac{1}{2(A+B+2)} \left(\frac{B \sqrt{A}}{\sqrt{A+ B +2}\left(1-\sqrt{\frac{A}{A + B + 2}}\right)} - A - 1 \right)
        + \log\left(1-\sqrt{\frac{A}{A + B + 2}}\right) \\
    &=& \frac{1}{2(A+B+2)} \left( \frac{B \sqrt{A}}{\sqrt{A + B +2} -\sqrt{A}} - A - 1 \right)
        + \log\left(1-\sqrt{\frac{A}{A + B + 2}}\right) \\
    &\le& \frac{1}{2(A+B+2)} \left( \frac{\sqrt{B} / 3}{\sqrt{A + B +2} -\sqrt{A}} - A - 1 \right) -\sqrt{\frac{A}{A + B + 2}} \\
    &\le& \frac{1}{2(A+B+2)} \left( \frac{1}{3} \sqrt{\frac{B}{B+2}} - A - 1 \right) -\sqrt{\frac{A}{A + B + 2}} \\
    &\le& - \frac{A}{2(A+B+2)} - \sqrt{\frac{A}{A + B + 2}} \\
    &\le& 0
\end{eqnarray*}
and similarly,
\begin{eqnarray*}
\frac{\partial f_2}{\partial B} &=&
    - A \left(\frac{\sqrt{\frac{B}{A+B+2}}}{2B} + \frac{1}{2(A+B+2)} \right)
    + \log\left(\sqrt{\frac{B}{A + B + 2}} \right)
    + B \left( \frac{A + 2}{2B(A+B+2)}\right)
    - \frac{1}{2(A + B +2)} \\
    &=& \frac{1}{2(A+B+2)} \left(A+ 2 - A - 1 - A\sqrt{\frac{A+B+2}{B}} \right)
        + \log\left(\sqrt{\frac{B}{A + B +2}} \right) \\
    &=& \frac{1}{2(A+B+2)} \left(1 - A\sqrt{\frac{A+B+2}{B}} \right)
        + \frac{1}{2}\log\left(\frac{B}{A + B +2} \right)
\end{eqnarray*}
Now we can look at each term to observe that $\frac{\partial^2 f_2}{\partial A \partial B} < 0$.
Therefore this expression $\frac{\partial f_2}{\partial B}$ is maximized over $A$ for $A=0$.
We now examine this expression:
\begin{eqnarray*}
\frac{\partial f_2}{\partial B} \big\vert_{A=0} =
    \frac{1}{2(B+2)} + \frac{1}{2}\log\left(\frac{B}{B +2} \right)
    \le \frac{1}{2} \left(\frac{1}{B+2} + \frac{B}{B+2} - 1 \right)
    \le 0
\end{eqnarray*}

Therefore, the expressions $f_i$ are minimized at at the largest possible $B = \frac{1}{9A}$ for any given $A$ over our region.
We will now write $g_i(A) := f_i(A, \frac{1}{9A})$ for this evalutation at the extremal boundary.
If we can show that $g_i(A) \ge 0$ for all $A \ge \frac{1}{3}$ and $i=1,2$ we will be done.

We will perform a similar argument to show that $g_i$ is monotone increasing,  $g'_i(A) \ge 0$ for all $A \ge \frac{1}{3}$.
\begin{eqnarray*}
    g_1(A) &=& \log(1 + A) + A \log\left( \sqrt{\frac{A}{A + \frac{1}{9A} + 2}} \right) + \frac{1}{9A} \log\left(1 - \sqrt{\frac{A}{A + \frac{1}{9A} + 2}} \right)
    \\&& + \frac{1}{2} \log(2 \pi) - \frac{1}{2}\log(A + \frac{1}{9A} + 2) \\
    &=& \log(1 + A) + \frac{A}{2} \log(A) - \frac{1}{2}(1+A) \log(A + \frac{1}{9A} + 2) \\
    &&+ \frac{1}{9A} \log\left(1 - \sqrt{\frac{A}{A + \frac{1}{9A} + 2}} \right)
    + \frac{1}{2} \log(2 \pi) \\
\end{eqnarray*}
Note that the function $p(A) = A + \frac{1}{9A}$ is increasing in $A$ for $A \ge \frac{1}{3}$.
We can conservatively bound $g$ from below noting $\frac{1}{9A} \le 1$ in our region.
\begin{eqnarray*}
    g_1(A) &\ge& = \log(1 + A) + \frac{A}{2} \log(A) - \frac{1}{2}(1+A) \log(A + 3)
    + \frac{1}{9A} \log\left(1 - \sqrt{\frac{A}{A + 2}} \right)
    + \frac{1}{2} \log(2 \pi) \\
    &\ge& \log(1 + A) + \frac{A}{2} \log(A) - \frac{1}{2}(1+A) \log(A + 3)
    - \frac{1}{9A} \sqrt{A} + \frac{1}{2} \log(2 \pi) =: \tilde{g}_1(A)
\end{eqnarray*}
Now we can use calculus to say that:
\begin{eqnarray*}
    \tilde{g}'_1(A) &=& \frac{1}{A+1} + \frac{1}{A+3} + \frac{\log(A)}{2}
    + \frac{1}{18 A^{3/2}} - \frac{1}{2}\log(A+3) \\
    &\ge& \frac{1}{A+1} + \frac{1}{A+3} + \frac{1}{18 A^{3/2}} + \frac{1}{2} \log(\frac{A}{A+3})
\end{eqnarray*}
This expression is monotone decreasing in $A$ and with a limit $\ge 0$ and so we can say that $\tilde{g}_1(A)$ is monotone increasing.
Therefore $g_1(A) \ge \tilde{g}_1(A) \ge \tilde{g}_1(1/3)$ for all $A$.
We can explicitly evaluate this numerically and $\tilde{g}_1(1/3) > 0.01$ so we are done.

The final piece of this proof is to do a similar argument for $g_2(A)$
\begin{eqnarray*}
    g_2(A) &=& \log(1 + A) + A \log\left( 1 - \sqrt{\frac{\frac{1}{9A}}{A + \frac{1}{9A} + 2}} \right) + \frac{1}{9A} \log\left(\sqrt{\frac{\frac{1}{9A}}{A + \frac{1}{9A} + 2}} \right) \\
    &&+ \frac{1}{2} \log(2 \pi) - \frac{1}{2}\log(A + \frac{1}{9A} + 2) \\
    &=& \log(1+A) + A \log\left( 1 - \sqrt{\frac{1}{9A^2 + 18A + 1}} \right) + \frac{1}{2} \left(\frac{1}{9A}\log\left(\frac{1}{9A} \right) \right) \\
    && - \frac{1}{2}\left(\frac{1}{9A} + 1\right)\log \left(A + \frac{1}{9A} + 2 \right) + \frac{1}{2}\log(2 \pi) \\
    &\ge& \log(1+A) + A \left(-\frac{1}{\sqrt{9A^2}}\right) + \frac{1}{2} \left(\frac{1}{9A}\log\left(\frac{1}{9A} \right) \right)
     - \frac{1}{2}\left(\frac{1}{3} + 1\right)\log \left(A + \frac{1}{3} + 2 \right) + \frac{1}{2}\log(2 \pi) \\
    &\ge& \log(1+A) - \frac{1}{3} - \frac{1}{2e} - \frac{2}{3}\log(A + \frac{7}{3}) + \frac{1}{2}\log(2 \pi) =: \tilde{g}_2(A)
\end{eqnarray*}
Now, once again we can see that $\tilde{g}_2$ is monotone increasing:
\begin{eqnarray*}
    \tilde{g}'_2(A) &=& \frac{1}{1+A} - \frac{2/3}{A+7/3} \\
    &=& \frac{A + 5}{(A+1)(3A + 7)} \ge 0
\end{eqnarray*}
We complete the argument by noting $g_2(A) \ge \tilde{g}_2(A) \ge \tilde{g}_2(1/3) > 0.01$, which concludes our proof of the PDF double crossing in this region.

\subsection{Recap}
Using the results of the previous sections we complete the proof of Lemma \ref{lem: Gauss Beta} for Gaussian vs Beta dominance for all possible $\alpha, \beta > 0$ such that $\alpha + \beta \ge 1$.
Piecing together Lemma \ref{le:DirchletToBeta} with Lemma \ref{lem: Gauss Beta} completes our proof of Lemma \ref{lem:Gaussian-Dirichlet}.
We imagine that there is a much more elegant and general proof method available for future work.

\end{document}